\documentclass[preprint,12pt]{elsarticle}

\usepackage{amsmath,amssymb,amsfonts}
\usepackage{algorithmic}
\usepackage{graphicx}
\usepackage{textcomp}
\usepackage{booktabs}
\usepackage{array}
\usepackage{tabularx}
\usepackage[T1]{fontenc}

\journal{}

\begin{document}

\begin{frontmatter}



\title{CPKD: Clinical Prior Knowledge-Constrained Diffusion Models for Surgical Phase Recognition in Endoscopic Submucosal Dissection}

\author[add1]{Xiangning Zhang\fnref{equal}}
\ead{zxnyyyyy@sjtu.edu.cn}
\author[add2]{Jinnan Chen\fnref{equal}}
\ead{23332@renji.com}
\author[add2]{Qingwei Zhang\fnref{equal}}
\ead{zhangqingwei@renji.com}
\author[add3]{Yaqi Wang}
\ead{wangyaqi@cuz.edu.cn}
\author[add4]{Chengfeng Zhou}
\ead{joe_chief@hnu.edu.cn}
\author[add2]{Xiaobo Li\corref{corresponding}}
\ead{lxb1969@sjtu.edu.cn}
\author[add1]{Dahong Qian\corref{corresponding}}
\ead{dahong.qian@sjtu.edu.cn}

\cortext[corresponding]{Corresponding author.}
\fntext[equal]{The three authors contribute equally to this work.}

\affiliation[add1]{organization={School of Biomedical Engineering},
            addressline={Shanghai Jiao Tong University}, 
            city={ShangHai},
            postcode={200000},
            country={China}}
\affiliation[add2]{organization={Division of Gastroenterology and Hepatology, Shanghai Institute of Digestive Disease, NHC Key Laboratory of Digestive Diseases, Renji Hospital,},
	addressline={Shanghai Jiao tong University School of Medicine}, 
	city={ShangHai},
	postcode={200000}, 
	country={China}}
\affiliation[add3]{organization={College of Media Engineering},
            addressline={Communication University of Zhejiang}, 
            city={Hangzhou},
            postcode={310018},
            country={China}}
\affiliation[add4]{organization={Aier Institute of Digital Ophthalmology and Visual Science},
            addressline={Changsha Aier Eye Hospital}, 
            city={Changsha},
            postcode={410000},
            country={China}}
 
\begin{abstract}
Gastrointestinal malignancies constitute a leading cause of cancer-related mortality worldwide, with advanced-stage prognosis remaining particularly dismal. Originating as a groundbreaking technique for early gastric cancer treatment, Endoscopic Submucosal Dissection has evolved into a versatile intervention for diverse gastrointestinal lesions. While computer-assisted systems significantly enhance procedural precision and safety in ESD, their clinical adoption faces a critical bottleneck: reliable surgical phase recognition within complex endoscopic workflows. Current state-of-the-art approaches predominantly rely on multi-stage refinement architectures that iteratively optimize temporal predictions. In this paper, we present Clinical Prior Knowledge-Constrained Diffusion (CPKD), a novel generative framework that reimagines phase recognition through denoising diffusion principles while preserving the core iterative refinement philosophy. This architecture progressively reconstructs phase sequences starting from random noise and conditioned on visual-temporal features. To better capture three domain-specific characteristics, including positional priors, boundary ambiguity, and relation dependency, we design a conditional masking strategy. Furthermore, we incorporate clinical prior knowledge into the model training to improve its ability to correct phase logical errors. Comprehensive evaluations on ESD820, Cholec80, and external multi-center demonstrate that our proposed CPKD achieves superior or comparable performance to state-of-the-art approaches, validating the effectiveness of diffusion-based generative paradigms for surgical phase recognition.
\end{abstract}


\begin{keyword}
Endoscopic submucosal dissection, Surgical phase recognition, Diffusion model, Clinical prior knowledge.
\end{keyword}

\end{frontmatter}



\section{Introduction}
\label{sec:introduction}
\begin{figure*}[!t]
	\centerline{\includegraphics[width=\textwidth]{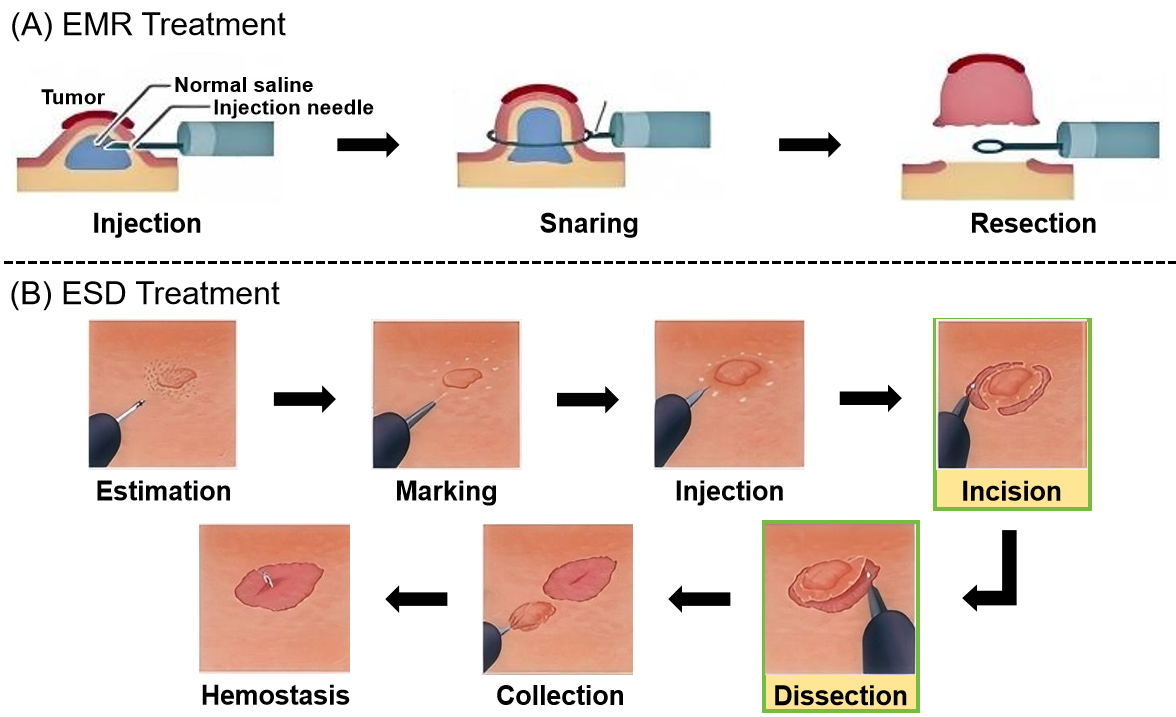}}
	\caption{(A) The treatment steps of EMR. (B) The treatment steps of ESD. The green border indicates steps specific to ESD}
	\label{background}
\end{figure*}

Gastrointestinal cancers represent a significant global health burden, accounting for more than one-third of cancer-related deaths \cite{huang2023updated, 2024nerve}. Endoscopic submucosal dissection (ESD), a technique involving seven standardized steps (as shown in Fig.\ref{background}) to achieve submucosal layer resection of gastrointestinal lesions, has transformed early-stage tumor management by enabling en bloc excision with minimal trauma, lower recurrence, and quicker recovery compared to endoscopic mucosal resection (EMR) \cite{mccarty2020endoscopic}. Although ESD has emerged as a groundbreaking minimally invasive technique for early gastric cancer treatment, its intricate procedural workflow still poses significant intraoperative risks and challenges in standardization \cite{maier2017surgical}. This underscores the urgent need for computer-assisted systems to enhance procedural safety and reproducibility. Within such systems, surgical phase recognition (SPR) serves as a cornerstone technology, as ESD’s multi-phase workflow exhibits phase-specific complication profiles. Accurate SPR enables real-time surgical monitoring, context-aware risk prediction, and workflow optimization \cite{huaulme2020offline}, while also facilitating structured video archiving for surgical education and competency assessment \cite{vercauteren2019cai4cai}. Consequently, automated SPR has become an essential component in advancing intelligent ESD assistance systems.

Surgical phase recognition has witnessed significant advancements in recent years with the proliferation of multi-stage architectures \cite{czempiel2020tecno, yi2022not, zhang2021swnet, gao2021trans, czempiel2021opera, zhang2024sprmamba, cao2024sr}. These models typically cascade multiple refinement stages: an initial prediction module followed by successive correction modules. Implementations range from TeCNO's dilated temporal convolutions \cite{czempiel2020tecno,2016Temporal,farha2019ms} to Opera's attention mechanisms \cite{czempiel2021opera,vaswani2017attention} and so on. While effective in capturing temporal dependencies and mitigating over-segmentation \cite{2023Deep}, such approaches exhibit two limitations: (1) \cite{czempiel2020tecno} reported that the parameter count of multi-stage TCNs is 2.3 times that of single-stage models, their complex multi-stage structures increase computational overhead during deployment. (2) the chained discriminative design lacks explicit mechanisms to incorporate clinical phase priors, resulting in compromised logical coherence.

To address these constraints, we introduce a novel generative paradigm that synergizes iterative refinement with diffusion modeling. Reinterpreting phase recognition as conditional sequence generation, our framework progressively reconstructs surgical phases through denoising diffusion \cite{2023Diffusion,dhariwal2021diffusion,ho2020denoising,song2020denoising}. The methodology leverages two complementary processes of the diffusion model: (1) a forward diffusion chain that systematically introduces noise to phase sequences, and (2) a learned reverse process that iteratively recovers the surgical phase from noisy inputs. Beyond learning discriminative frame-phase mappings, this formulation implicitly captures surgical phase distributions through generative modeling. We explicitly enhance three domain-specific characteristics. \textbf{Positional priors} exhibit a strong temporal localization pattern during the surgical phase. Taking a video of ESD surgery as an example, phases of estimation tend to appear at the beginning of the video, while clips are mostly located at the end.
\textbf{Boundary ambiguity} reflects that transitions between phases are visually gradual and thus lead to feature-space ambiguities around phase boundaries. \textbf{Relation dependency} represent surgical phases usually adhering to some intrinsic temporal ordering, e.g., clippingcutting typically follows calottriangledissection. This relation dependency differs from the position prior since it focuses on the arrangements relative to other phases. A unified adaptive conditional masking strategy dynamically modulates feature availability during training, forcing the model to internalize these priors. This differs fundamentally from multi-stage approaches by unifying temporal refinement and domain knowledge integration within a single generative architecture.

In this paper, our proposed Clinical Prior Knowledge-Constrained Diffusion (CPKD) framework reformulates surgical phase recognition as a conditional sequence generation task, where framewise phase labels are synthesized under the guidance of input video features. The architecture integrates two synergistic innovations. \textbf{Clinical Prior Knowledge Constraints (CPKC):} Explicitly encodes surgical phase logic through domain-specific rule embeddings derived from established procedural protocols. \textbf{Conditional Masking Strategy:} Dynamically modulates visual conditioning during training using stochastic masking ratios (0–100\%), forcing the model to reason about temporal positioning, phase duration, and procedural dependencies. This co-training paradigm creates synergistic effects: while CPKC directly injects clinical knowledge via constraint-based regularization, conditional masking enhances robustness against real-world observational uncertainties. Crucially, both mechanisms operate exclusively during training, enabling the diffusion backbone to internalize. During inference, phase predictions are generated through 8-step iterative denoising of Gaussian noise, conditioned solely on visual inputs. This co-design ensures that clinical expertise and data-driven adaptation are seamlessly integrated, addressing both logical consistency and generalization challenges in surgical phase analysis to overcome the limitations of current multi-stage approaches.

The effectiveness of our method is demonstrated by the experiments on two datasets, ESD820 \cite{zhang2024sprmamba, chen2025renji} and Cholec80 \cite{twinanda2016endonet}, on which our model performs better or on par compared to state-of-the-art methods. In summary, our contributions are threefold:
\begin{itemize}
  \item We formulate surgical phase recognition as a conditional generation task and propose a new iterative refinement framework based on the denoising diffusion process. Unlike traditional discriminative models that only learn frame-stage mappings, this study uses the bidirectional process of diffusion models (forward noise addition + backward denoising) to implicitly model the temporal distribution of surgical stages, providing a new paradigm for dynamic process modeling.
  \item We design a conditional masking strategy including global masking, phase transition making, and phase relation masking to further utilize the prior surgical phase.
  \item 
  We follow the procedure established by professional endoscopists to propose a Clinical Prior Knowledge-Constraint (CPKC) unit. Unlike existing studies, which mostly use weak supervision \cite{jin2017sv} or graph networks \cite{kadkhodamohammadi2022patg} to indirectly capture phase relationships, we explicitly encode clinical rules such as ‘marking must precede injection’ using LTL logic formulas to achieve interpretable control of logical constraints.
\end{itemize}

\section{Related Work}
\label{sec:related work}
\subsection{Surgical Phase Recognition}
Surgical phase recognition \cite{2023Deep} takes a set of surgical video frames as input and predicts the phase category for each frame. To model both short and long-term dependencies between phases, a variety of temporal models have been used in the literature. From the early methods use of sliding windows to model changes in visual appearance over a short period, capturing long-term dependencies from this short-term information through sequential models such as Hidden Markov Models \cite{twinanda2016endonet} and Recurrent Networks \cite{twinanda2017vision, jin2017sv}, to Temporal Convolutional Networks (TCNs) \cite{czempiel2020tecno, yi2022not, zhang2021swnet} that can efficiently model temporal dependencies across different time spans through their flexible sensory fields, to the Transformer \cite{gao2021trans, czempiel2021opera, liu2023skit} which significantly improves the performance of the model in complex surgical scenarios through a self-attention mechanism, and more recently Mamba \cite{zhang2024sprmamba, cao2024sr} which features transformer-level performance and linear complexity through the introduction of a data-dependent SSM layer and a selection mechanism using parallel scanning. Multi-stage models \cite{czempiel2020tecno, yi2022not, zhang2021swnet, gao2021trans, czempiel2021opera, zhang2024sprmamba, cao2024sr} are particularly noteworthy in that they capture temporal context and reduce over-segmentation errors through progressive optimization \cite{2023Deep}. In addition to architectural design, another series of studies has focused on improving model performance and generalization through multi-task learning \cite{twinanda2016endonet, JIN2020101572, 2021Multi, tao2023last} or multi-modal fusion approaches \cite{2022Visual}.

For surgical phase recognition, the position prior, the relational prior, the boundary ambiguity, and the clinical knowledge are four useful types of prior knowledge. The clinical prior, relational prior, and clinical knowledge have attracted the attention of researchers, while position prior has been less explored. To cope with the clinical prior, Jin et al. \cite{jin2017sv} designed a PKI scheme to enhance the consistency of phase prediction. Fujii et al. \cite{fujii2024egosurgery} introduced eye-movement information as an empirical semantic richness that guides the masking process and promotes better attention to semantically rich spatial regions. Malayeri et al. \cite{malayeri2025arthrophase} proposed an effective mechanism, the Surgical Progress Index, which implicitly integrates the global temporal context and provides a valuable measure of surgical progress. For the relational prior, PATG \cite{kadkhodamohammadi2022patg} utilizes graph neural networks to facilitate paragraph-level relational inference. For contextual relations between neighboring phases, Chen et al. \cite{chen2025text} employed multi-modal learning using textual cues as supervision. For boundary ambiguity, Liu et al. \cite{liu2025lovit} propose a phase transition-aware supervised mechanism to emphasize critical transition moments in surgery. In this paper, we design a CPKC unit to improve the robustness and accuracy of the model by applying phase logic constraints to the model and simply using conditional masking strategies to deal with position prior, boundary ambiguity, and relational prior simultaneously.

One related work \cite{ding2023less} also employs generative learning for surgical phase recognition. The method generates plausible pseudo-labels from timestamp annotations via uncertainty estimation and introduces loop training to train the model from the generated pseudo-labels in an iterative manner. In contrast, our approach generates surgical phase label sequences through the diffusion model.

\subsection{Diffusion models}
Diffusion models \cite{2023Diffusion, ho2020denoising, song2020denoising}, theoretically unified with the score-based generative approaches \cite{song2019generative, 2020Improved}, have gained prominence for their stable training dynamics and non-adversarial learning paradigm. These models have demonstrated remarkable success across multiple domains: (1) image generation \cite{dhariwal2021diffusion, rombach2022high}, (2) natural language generation \cite{yu2022latent}, (3) cross-modal synthesis \cite{2021Vector, kim2022diffusionclip}, (4) audio generation \cite{2022BDDM, leng2022binauralgrad}, and so on. Recent theoretical advances \cite{dinh2023pixelasparam} have further enhanced sampling efficiency through gradient-based guidance. While computer vision applications have begun emerging in detection \cite{chen2023diffusiondet} and segmentation \cite{baranchuk2021label, 2021SegDiff}, video-related implementations remain scarce, limited primarily to forecasting and gap-filling tasks \cite{hoppe2022diffusion, 2023Generation}. Notable exceptions include video memorability prediction \cite{sweeney2022diffusing} and caption generation \cite{zhong2023refined}. Our work pioneers the adaptation of diffusion models' iterative refinement capability to surgical phase recognition, representing the first application in surgical video analysis to our knowledge.

\section{Preliminaries}
\label{sec:preliminaries}
In this section, we provide a concise overview of the principles underlying diffusion models, which form the basis of our proposed methodology. Diffusion models \cite{ho2020denoising, song2020denoising} are a class of generative models that learn to generate data by simulating a gradual denoising process. The framework consists of two key phases: the forward process and the reverse process. During the forward process, Gaussian noise is added in a series of steps to corrupt the original data into noise data. In the reverse process, a neural network is trained to iteratively denoise the data, effectively reversing the corruption and recovering the original data distribution.

\textbf{The forward process} transforms the original data $x_{0}$ into noise data $x_{t}$:
\begin{equation}x_{t}=\sqrt{\lambda(t)}x_{0}+\sqrt{1-\lambda(t)}\epsilon.
\label{eq1}
\end{equation}
The noise $\epsilon \sim \mathcal{N}(0, \mathbf{I})$ is sampled from a standard normal distribution, with its intensity defined by a decreasing function $\lambda(t)$ of the timestep $t \in \{1,2,\dots,T\}$, where T represents the total number of timesteps, and $\lambda(t)$ determines the strength of the noise added to the original data $x_{0}$ according to a predefined variance schedule \cite{ho2020denoising}.

\textbf{The reverse process} start from noise data $x_{T}$ and progressively removes noise to recover original data $x_{0}$, with one step in the process defined as $p_{\theta}(x_{t-1}|x_{t})$:
\begin{equation}p_{\theta}(x_{t-1}|x_{t})=\mathcal{N}(x_{t-1};\pi_{\theta}(x_{t},t),\gamma^{2}_{t}\mathbf{I}).
\label{eq2}
\end{equation}
where $\pi_{\theta}$ and $\gamma_{t}$ are predicted mean and co-variance, respectively, derived from a step-dependent neural network.

There exist multiple approaches \cite{2022Understanding} to parameterize \( p_\theta \), such as predicting the mean as shown in Eq.\eqref{eq2}, directly predicting the noise \( \epsilon \), or predicting the original data \( x_0 \). In our work, we adopt the \( x_0 \)-prediction approach, where a neural network \( f_\theta(x_t, t) \) is employed to estimate \( x_0 \) rather than directly predicting \( \mu_\theta(x_t, t) \) as in Equation 2. To train the model, a mean squared error loss is applied to minimize the discrepancy between the predicted \( f_\theta(x_t, t) \) and the ground truth \( x_0 \): \( \mathcal{L} = \|f_\theta(x_t, t) - x_0\|^2 \), where the timestep \( t \) is randomly sampled from \( \{1, 2, \dots, T\} \) during each training iteration.

During inference, the generation process begins with a pure noise sample \( x_T \sim \mathcal{N}(0, \mathbf{I}) \). The model then iteratively denoises the sample by following the update rule \cite{song2020denoising} below, leveraging the trained neural network \( f_\theta \):
\begin{equation}x_{t-1}=\sqrt{\lambda(t-1)}f_\theta(x_t, t)+\sqrt{1-\lambda(t-1)-\gamma^{2}_{t}}\frac{x_t-\sqrt{\lambda(t)}f_\theta(x_t, t)}{\sqrt{1-\lambda(t)}}+\gamma_t\epsilon.
\label{eq3}
\end{equation}

By iteratively applying the update rule in Eq.\eqref{eq3}, a new sample \( x_0 \) can be generated from the learned distribution \( p_\theta \) through a trajectory \( x_T, x_{T-1}, \dots, x_0 \). To enhance computational efficiency, some advanced sampling strategies skip intermediate steps in the trajectory, generating samples along a reduced sequence such as \( x_T, x_{T-\Delta}, \dots, x_0 \), as proposed in \cite{song2020denoising}. This approach significantly accelerates the sampling process while maintaining sample quality.

Conditional information can be incorporated into the diffusion model to guide the generation process. The conditional model is expressed as \( f_\theta(x_t, t, C) \), where \( C \) represents the additional conditional input. In existing research, common forms of conditional information include class labels \cite{dhariwal2021diffusion}, text prompts \cite{2021Vector,kim2021diffusionclip}, and image guidance \cite{preechakul2022diffusion}. These conditioning mechanisms enable precise control over the generated outputs, enhancing the model's versatility and applicability to various tasks.

\section{Methodology}
\label{sec:methodology}
\begin{figure*}[!t]
	\centerline{\includegraphics[width=\textwidth]{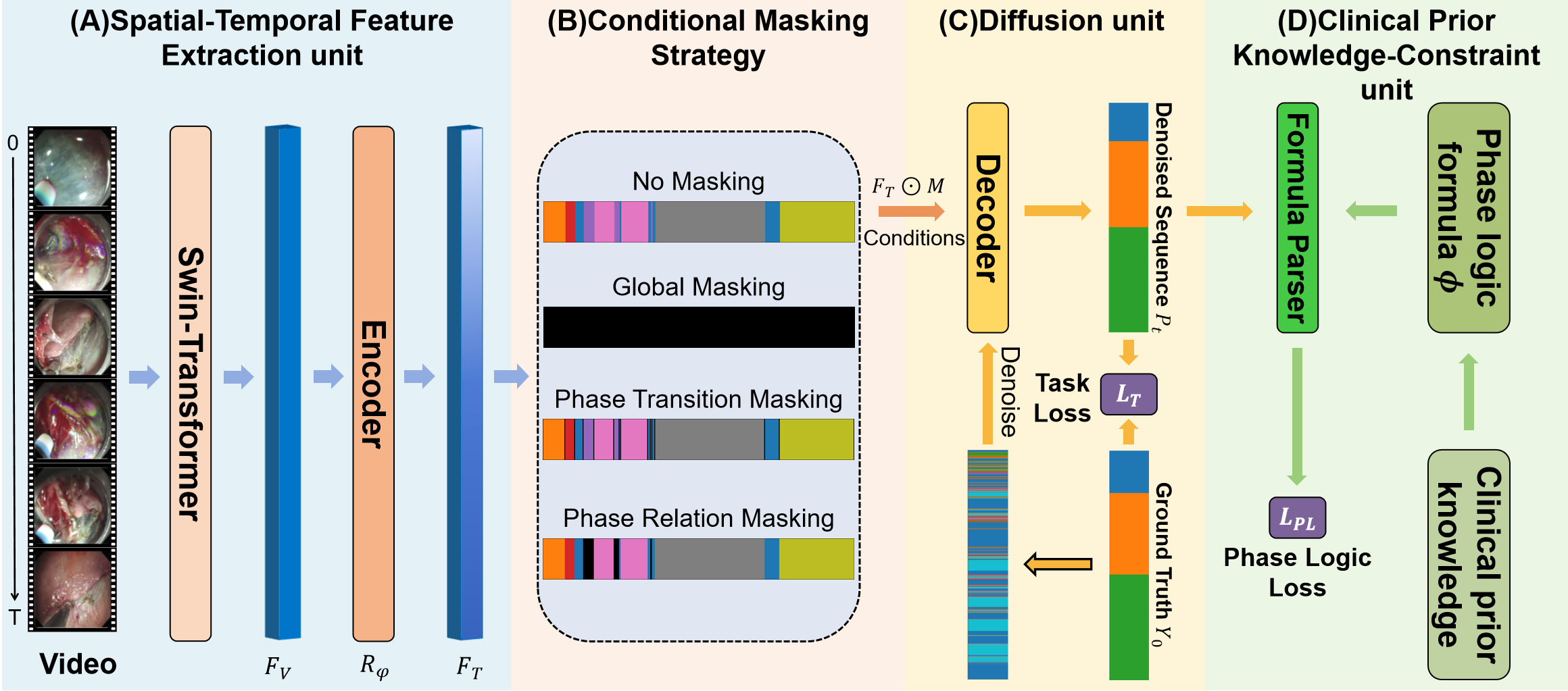}}
	\caption{A schematic overview of the proposed CPKD architecture during training. (A) shows the Spatial-temporal feature extraction unit. Video frames are processed through a Swin-transformer to extract spatial features \(F_v\), and then through an encoder \(R_\varphi\) to generate temporal features \(F_T\). (B) Illustration of the conditional masking strategy. The black locations are masked. Other different colors represent different phases. (C) shows the Diffusion unit. Given spatial-temporal features as conditional information, the model learns the mapping between noise sequences and phase labels. (D) shows the CPKC unit. A set of clinical prior knowledge on surgical phases is converted into a phase logic formula. A formula evaluator takes \( \phi \) and \( P_t\) as input, and evaluates the consistency between the two, resulting in a logic loss \( L_{PL}\).}
	\label{training}
\end{figure*}
In this work, we formulate the ESD surgical phases recognition as a conditional generation problem of ESD surgical phase sequences. Given the D-dimensional input features \( F_v \in \mathbb{R}^{T \times D} \) of a video with \( T \) frames, we approximate the data distribution of the ground truth phase sequence \( Y_0 \in \{0, 1\}^{T \times C} \) with \( C \) classes of phases.

\subsection{Overview}
As shown in Fig.\ref{training}, our Clinical Prior Knowledge-Constrained Diffusion framework comprises three key components: a Spatial-Temporal Feature Extraction (STFE) unit, a Diffusion unit, and a Clinical Prior Knowledge-Constraint (CPKC) unit. The STFE unit first encodes surgical videos into a sequence of embedded features through Swin-transformer \cite{2021Swin}, which are then processed by an encoder \( R_\varphi \) to capture long-range temporal dependencies and generate task-oriented features. These enhanced features serve as input to the diffusion unit. During training, the diffusion unit learns to reconstruct the original surgical phase annotations through a denoising process, where it progressively recovers the ground truth from artificially corrupted phase sequences (with injected noise) conditioned on the visual features \( F_T \). To further improve temporal coherence and incorporate domain-specific knowledge, we introduce two novel mechanisms: 1) the conditional masking strategy that guides the model to capture phase priors, and 2) the CPKC unit, which explicitly enforces clinical prior knowledge to minimize logical inconsistencies in the generated sequences. At inference time, as shown in Fig.\ref{inference}, the unit iteratively refines a randomly initialized sequence via the reverse diffusion process to produce accurate surgical phase recognition. In the following sections, we will elaborate details of each unit.

\subsection{The STFE Unit}
For an input video \( V \) comprising \( T \) frames, denoted as \( V = \{v_1, \dots, v_T\} \), each frame at time stamp \( i \) is represented as \( V_i \in \mathbb{R}^{3 \times W \times H} \), where \( W \) and \( H \) are the width and height of the frame, respectively, and \( 3 \) corresponds to the number of color channels. To process the video data, we adopt Swin-B \cite{2021Swin} as a backbone network, which converts each frame into a fixed-length 1D feature vector. This transformation is formally expressed as:
\begin{equation}
\bar{F}_{V_i} = \text{Swin}(V_i).
\label{eq4}
\end{equation}
where \( \bar{F}_{V_i} \in \mathbb{R}^{1024} \) represents the extracted \( 1 \times 1024 \) feature vector for the \( i \)-th frame.

Subsequently, we input \( T\) length spatial features \( F_{V_i}\) into the encoder and convert it into a fixed length 2D feature vector. Encoder \( R_\varphi \) is a re-implemented ASFormer encoder \cite{asformer} as shown in Fig.\ref{fig2}, this transformation is expressed as:
\begin{equation}
\bar{F}_{T} = R_\varphi(F_V).
\label{FT}
\end{equation}
where \( \bar{F}_{T} \in \mathbb{R}^{T \times 256} \) represents the extracted \( T \times 256 \) feature vector for the \( T\)-length video.
 
\subsection{The Diffusion Unit}
\begin{figure*}[!t]
	\centerline{\includegraphics[width=0.7\textwidth]{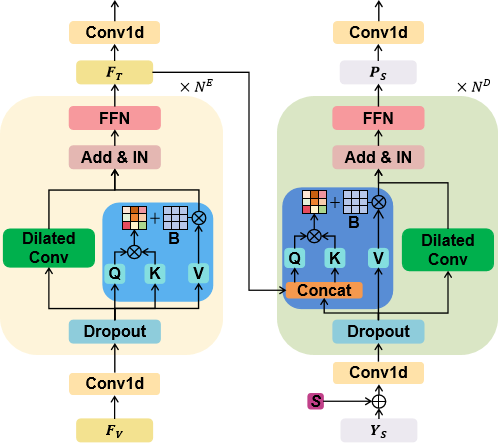}}
	\caption{A schematic overview of the encoder and decoder layer.}
	\label{fig2}
\end{figure*}
The diffusion unit establishes a diffusion process \( Y_0, Y_1, \dots, Y_T \) during training, which gradually transforms the ground truth phase sequence \( Y_0 \) into a nearly pure noise \( Y_T \). At inference, it performs a denoising process \( \hat{Y}_T, \hat{Y}_{T-1}, \dots, \hat{Y}_0 \), starting from a pure noise \( \hat{Y}_T \) and progressively recovering the predicted phase sequence \( \hat{Y}_0 \). This framework leverages the diffusion and denoising mechanisms to model the distribution of phase sequences effectively.

\textbf{Training.} To learn the underlying distribution of surgical phases, the model is trained to reconstruct the ground truth phase sequence from its corrupted versions. Specifically, at each training iteration, a random diffusion step \( t \in \{1, 2, \dots, T\} \) is selected. The ground truth sequence \( Y_0 \) is then corrupted by adding noise according to the predefined cumulative noise schedule, as described in Eq.\eqref{eq1}, resulting in a noisy sequence $Y_{T} \in [0, 1]^{T \times C}$:
\begin{equation}
Y_{T}=\sqrt{\lambda(t)}Y_{0}+\sqrt{1-\lambda(t)}\epsilon.
\label{eq5}
\end{equation}
where noise $\epsilon$ is from a Gaussian distribution.

Taking the corrupted sequence \( Y_t \) as input, a decoder \( D_\varphi \) is designed to denoise the sequence:
\begin{equation}
P_t = D_\varphi(Y_t, t, F_v \odot M).
\label{eq6}
\end{equation}
where the resulting denoised sequence \( P_t \in [0, 1]^{T \times C} \) represents the phase probabilities for each frame. In addition to \( Y_t \), the decoder incorporates two other inputs: the diffusion step \( t \) and the spatial-temporal features \( F_T \). The inclusion of \( t \) makes the model step-aware, enabling it to handle different noise intensities using the same architecture. The spatial-temporal features \( F_T \) serve as conditional information, ensuring that the generated phase sequences are not only broadly plausible but also consistent with the input video. Furthermore, the conditional information is enriched by element-wise multiplication of \( F_T \) with the conditional masking strategy \( M \), which significantly improve the accuracy, robustness, and interpretability of the model. The decoder \( D_\varphi \) is a modified version of the ASFormer decoder adapted to be step-dependent as shown in Fig.\ref{fig2}. This modification incorporates a step embedding into the input, following the approach described in \cite{ho2020denoising}, to enable the model to handle different noise levels effectively during the denoising process.

\begin{figure*}[!t]
	\centerline{\includegraphics[width=\textwidth]{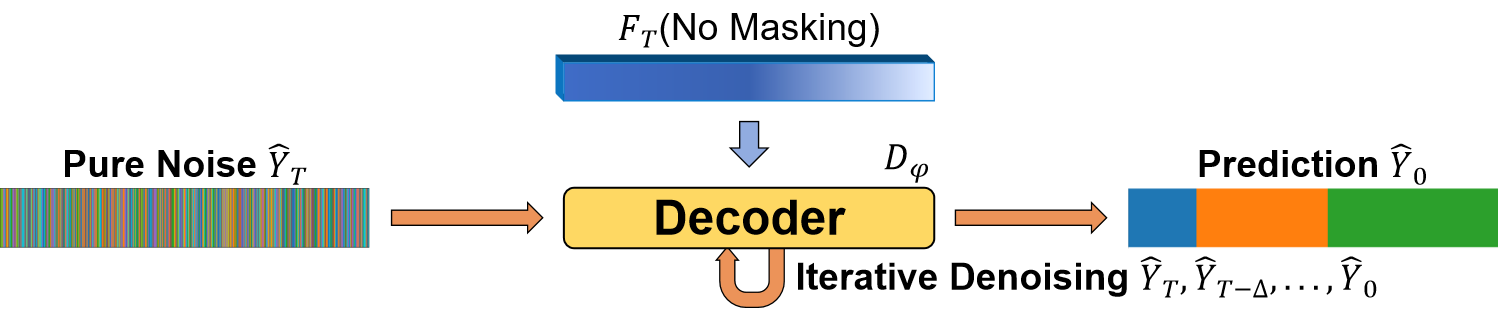}}
	\caption{A schematic overview of the proposed CPKD architecture during inference. At inference, the model begins with a random noise sequence and obtains results via an iterative denoising process.}
	\label{inference}
\end{figure*}
\textbf{Inference.} Intuitively, the denoising decoder \( D_\varphi \) is trained to adapt to sequences with arbitrary noise levels, enabling it to handle even a sequence composed entirely of random noise. As a result, the denoising decoder begins with a pure noise sequence \( \hat{Y}_T \sim \mathcal{N}(0, \mathbf{I}) \) and progressively reduces the noise during inference. Specifically, the update rule for each step is adapted from Eq.\eqref{eq3} as follows:
\begin{equation}
\hat{Y}_{t-1}=\sqrt{\lambda(t-1)}P_t+\sqrt{1-\lambda(t-1)-\gamma^{2}_{t}}\frac{\hat{Y}_{t}-\sqrt{\lambda(t)}P_t}{\sqrt{1-\lambda(t)}}+\gamma_t\epsilon.
\label{eq7}
\end{equation}
where \( \hat{Y}_{t-1} \) is passed into the decoder to obtain the next denoised sequence \( P_{t-1} \). This iterative refinement process generates a series of phase sequences \( \hat{Y}_T, \hat{Y}_{T-1}, \dots, \hat{Y}_0 \), culminating in \( \hat{Y}_0 \), which closely approximates the underlying ground truth and is regarded as the final prediction. To accelerate inference, our method employs a sampling trajectory \cite{song2020denoising} with skipped steps, denoted as \( \hat{Y}_T, \hat{Y}_{T-\Delta}, \dots, \hat{Y}_0 \). Notably, during inference, the encoded features \( F_T \) are fed into the decoder without masking. Additionally, we fix the random seed for the denoising process to ensure deterministic results in practice.

\subsection{Conditional Masking Strategy}
As shown in Fig.\ref{training}(B), we introduce three different conditional masking strategies: Global Masking Strategy, Phase Transition Masking Strategy and Phase Relation Masking Strategy.

\textbf{No Masking Strategy.} The first strategy is a basic all-one mask \( M_N = \mathbf{1} \), which allows all features to pass through to the decoder. This naive mask provides complete conditional information, enabling the model to fully utilize spatial features for mapping to phase classes.

\textbf{Global Masking Strategy.} The second strategy employs a full mask (\(M_{G}=\mathbf{0}\)), which entirely obscures visual features. In this setting, the decoder must rely exclusively on the sequential order and timestamps of phase labels as its sole conditioning signal. This forces the model to implicitly learn transition patterns and dependencies between surgical phases, thereby enhancing its robustness to noisy or missing visual data while improving sensitivity to temporal context.

\textbf{Phase Transition Masking Strategy.} Due to the inherent ambiguity of phase transitions, spatial features near boundaries may lack reliability. To mitigate this, the phase transition masking strategy \( M_T \) selectively removes features adjacent to transition boundaries based on the soft ground truth probabilities \( \bar{P} \). The mask is formally defined as \( M^T_i = \mathbf{1}(\bar{B}_i < 0.5) \) for each frame \( i \in \{1, 2, \dots, T\} \). By applying \( M_T \), the decoder is forced to rely less on potentially noisy boundary features and instead leverage richer contextual cues, such as the temporal structure of preceding and subsequent phases, as well as their durations. This approach enhances the model's robustness and generalization capability by prioritizing globally consistent phase semantics over local, transition-related artifacts.

\textbf{Phase Relation Masking Strategy.} Recognizing that surgical phases follow inherent ordinal relationships, we propose a phase relation masking strategy \( M_R \) to explicitly model these dependencies. During training, this approach randomly selects a phase class \( \tilde{c} \in \{1, 2, \dots, C\} \) and masks all corresponding segments using \( M^R_i = \mathbf{1}(Y_{0,i,\tilde{c}} \neq 1) \) for frames \( i \in \{1, 2, \dots, T\} \). By forcing the model to reconstruct masked phases solely from contextual information of adjacent phases, we effectively transform phase recognition into a conditional sequence completion task that mimics surgical reasoning. This strategy not only captures the procedural logic of surgical phases but also enhances the model's robustness to phase transitions by learning relational patterns rather than relying on absolute features. The resulting representation better reflects how surgeons naturally infer phases through understanding of procedural context and temporal dependencies

During training, four mask types are randomly selected: No mask \( M_N \), Global mask \( M_G \), Relation mask \( M_R \), and Phase transition mask \( M_T \). During inference, use No mask \( M_N \).

\subsection{The CPKC Unit}
\begin{figure*}[!t]
	\centerline{\includegraphics[width=\textwidth]{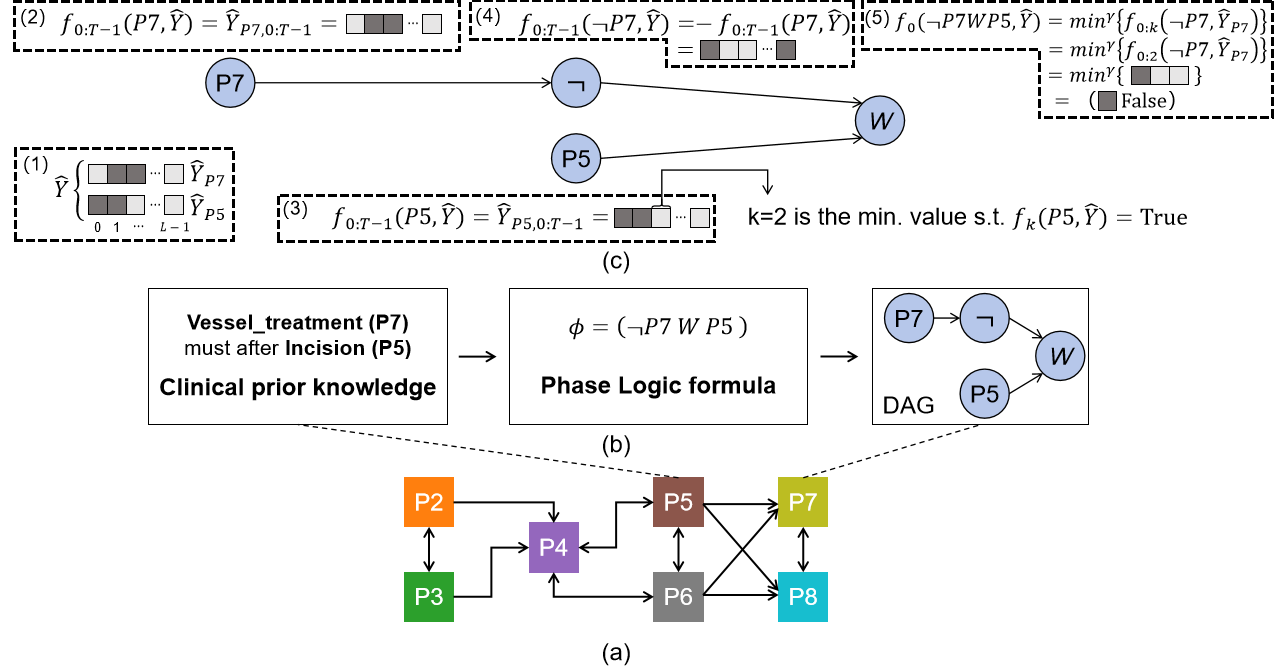}}
	\caption{(a) The order information of the phases defined in the ESD820 dataset. (b) The evaluation of an example constraint on two phases over L time steps. The constraint is first written as formula \( \phi \) = \( (\lnot P7WP5) \), then represented as a DAG. (c) The computation process of \( f_0(\lnot P7WP5, P_t)\) is illustrated through steps (1) - (5). (1) shows the phase predictions \(P_t\) are generated by the task model, with truth values represented visually (dark grey: True; grey: False). (2) and (3) show the evaluation of leaf nodes P7 and P5, which is the start of evaluation. (4) processes the evaluation output for P7 from (2) based on Eq.\eqref{eq:neg}. In (5), the temporal logic operator W combines \(\lnot P7\) and P5 following Eq.\eqref{eq14} to produce the complete formula evaluation }
	\label{CPKC}
\end{figure*}
We have observed that, unlike natural videos, the structure and sequence of ESD surgical videos are more defined, as endoscopists are required to strictly adhere to specified workflows and operational protocols as shown in Fig.\ref{CPKC}(a). To better leverage this characteristic and incorporate clinical prior knowledge, we engaged in in-depth discussions with an interdisciplinary team (comprising professional endoscopists). Based on these discussions, we have summarized three primary phase logic constraint rules in ESD surgery:
\begin{itemize}
  \item Marking and Estimation must be completed before Injection, Incision, ESD, Vessel\_treatment, and Clips;
  \item Injection, Incision, and ESD must be completed before Vessel\_treatment and Clips; 
  \item In the ESD surgical video, when one of the four stages of marking, injection, infection, and ESD appears, the other three stages must appear.
\end{itemize}

Based on these clinical rules, we have proposed a Clinical Prior Knowledge-Constraint Unit. Our ultimate goal is to integrate these constraints into the ESD surgical phase recognition model, so that the predicted results \( P_t \) not only align with the true labels \( Y_0 \), but also satisfy the aforementioned logical constraints. To achieve this, we have introduced a phase logic representation \( \Phi \) and a formula parser  \( f \), which enforces the constraint rules in \( \Phi \) through the output \( P_t \). Specifically, the parser \( f \) adjusts the predicted results according to the formula \( \phi \in \Phi \), ensuring they conform to the logical sequence of surgical phases.

\textbf{Formulae Syntax.} The definition of \( \Phi \) extends Linear Temporal Logic (LTL) \cite{pnueli1977temporal}. A formula \( \phi \in \Phi \) can take any of the following forms, separated by the "|" symbol:
\begin{equation}
\phi = \text{True} \ | \ \text{False} \ | \ p \ | \lnot\phi_1 | (\phi_1\vee\phi_2) | (\phi_1\wedge\phi_2) | \ X\phi_1 \ | \ F\phi_1 \ | \ (\phi_1 \, W \, \phi_2) \ | \ (\phi_1 \, S \, \phi_2).
\label{eq9}
\end{equation}
where \( p \in P \) represents an atomic proposition, and \( \phi_1, \phi_2 \in \Phi \). The modal operators \( X \) (\textbf{NEXT}), \( F \) (\textbf{EVENTUAL}), \( W \) (\textbf{WEAK\_UNTIL}), and \( S \) (\textbf{SINCE}) establish temporal relationships between propositions. In our framework, atomic propositions \( p \) correspond to specific phases. Notably, the definition in Eq.\eqref{eq9} is recursive. For instance, if \( p_1 \in \Phi \), then \( X \dots X F p_1 \in \Phi \). Semantics: \( X\phi \) is satisfied if \( \phi \) holds in the next time step. \( F\phi \) is satisfied if \( \phi \) eventually holds at some point in the sequence. \( \phi_1 \, W \, \phi_2 \) is satisfied if \( \phi_1 \) holds continuously until \( \phi_2 \) becomes true (with no requirement that \( \phi_2 \) must eventually hold). \( \phi_1 \, S \, \phi_2 \) is satisfied if \( \phi_1 \) holds continuously after \( \phi_2 \) has been true.

\textbf{Formula Parser.} A formula is considered satisfied by a truth value assignment if the assignment semantically adheres to the specified constraints. In this work, each atomic proposition \( p_i \) is associated with a predicted value \( P_{t_i} \). The satisfiability of a formula \( \phi \) is determined through an formula parser, which takes \( \phi \) and \( P_t \) as inputs:
\begin{equation}
    f_t(\phi, P_t) = f(\phi, P_{p_1:p_N, t:T-1}) : \Phi \times \mathbb{R}^{T-t} \times \cdots \times \mathbb{R}^{T-t} \rightarrow \mathbb{R}.
\label{eq10}
\end{equation}
where \( \times \) denotes the Cartesian product. The function \( f \) evaluates two arguments: the formula \( \phi \) and the model predictions \( P_{p_1:p_N} \). The parameter \( t \in [0, T-1] \) specifies the starting time for the evaluation. For instance, \( t = 4 \) indicates that the evaluation is performed between \( \phi \) and the predictions starting from the five frame, \(i.e., P_{p_1:p_N, 4:T-1} \). The output of \( f \) is a satisfaction score that quantifies the consistency between the predictions \( P_s \) and the formula \( \phi \). 

Eq.\eqref{eq10} is abstract and requires detailed definitions for all possible forms of a formula \( \phi \) as outlined in Eq.\eqref{eq9}. Our goal is to expand the definition to ensure that \( \Phi \) is logically sound, meaning there is a clear relationship between \( f_t(\phi, P_t) \) and the satisfaction of \( \phi \) given \( P_t \). Specifically, we aim for \( f_t(\phi, P_t) > 0 \) to imply that \( \phi \) is satisfied by \( P_t \) at time \( t \). Additionally, \( \Phi \) must be differentiable to integrate seamlessly with task models. To achieve these objectives, we first define the evaluation for constants and atomic propositions:

\begin{equation}
f_t(\text{True}, P_t) = +\infty, \quad f_t(\text{False}, P_t) = -\infty, \quad f_t(p, P_t) = P_{p,t}.
\label{eq11}
\end{equation}
Here, the evaluation results for True and False are always positive and negative, respectively, because True is always satisfied, while False is never satisfied. For an atomic proposition \( p \), the formula \( \phi = p \) is satisfied at time \( t \) if \( P_{p,t} > 0 \), indicating that phase \( p \) occurs at time \( t \). 

Next, we define the evaluation for operators ``$\neg$'' (negation), ``$\vee$'' (logical or) , and ``$\wedge$'' (logical and):
\begin{align}
f_{t}(\neg\phi_{1},P_t) &= -f_{t}(\phi_{1},P_t), \label{eq:neg} \\
f_{t}(\phi_{1} \vee \phi_{2}, P_t) &= \max^{\gamma}\{f_{t}(\phi_{1}, P_t), f_{t}(\phi_{2}, P_t)\}. \label{eq:or} \\
f_{t}(\phi_{1} \wedge \phi_{2}, P_t) &= \min^{\gamma}\{f_{t}(\phi_{1}, P_t), f_{t}(\phi_{2}, P_t)\}, \label{eq:and} 
\end{align}
where $\gamma$ is a smoothing parameter of function \(\min^\gamma\{p_{1:T-1}\} = -\gamma\log\sum_{i=1}^{T-1}e^{-\frac{p_{i}}{\gamma}}\) greater than zero, which approximates the minimum value of $\{p_{1:T-1}\}$ \cite{xu2022don}, and \(\max^{\gamma}\{p_{1:T-1}\} = -\min^{\gamma}\{-p_{1:T-1}\}\). It can be shown \cite{xu2022don} that \(\lim_{\gamma\to\infty}\min^{\gamma}\{p_{1:T-1}\} = \min\{p_{1:T-1}\}\). In Eqn.~\eqref{eq:neg}, the operator $\neg$ flips the sign of $f_{t}(\phi_{1},P_t)$, reflecting the negation semantics of $\neg$. In Eqn.\eqref{eq:or}, $\max^{\gamma}\{f_{t}(\phi_{1},P_t),f_{t}(\phi_{2},P_t)\}$ will be negative (False) if both $\phi_{1}$ and $\phi_{2}$ are False, and will be positive (True) if $\phi_{1}$ or $\phi_{2}$ or both are True. This is consistent with the semantics of $\vee$. The same rationale applies to $\min^{\gamma}$ for $\wedge$ in Eqn.\eqref{eq:and}.

Finally, we define the modal operators \( X \), \( F \), \( W \), and \( S \), which are unique to \( \Phi \). Intuitively, evaluating \( X\phi_1 \) is equivalent to evaluating \( \phi_1 \) at the next time step. \( F\phi_1 \) requires that \( \phi_1 \) is satisfied at least once in the future. \( \phi_1 \, W \, \phi_2 \) and \( \phi_1 \, S \, \phi_2 \) require \( \phi_1 \) to be continuously satisfied during the time period specified by \( \phi_2 \). Formally, these operators are defined as follows:
\begin{equation}
f_t(X\phi_1, P_t) = f_{t+1}(\phi_1, P_t).
\label{eq12}
\end{equation}
\begin{equation}
f_t(F\phi_1, P_t) = \max^\gamma \left\{ f_{t:T-1}(\phi_1, P_t) \right\}.
\label{eq13}
\end{equation}
\begin{equation}
f_t(\phi_1 \, W \, \phi_2, P_t) = \min^\gamma \left\{ f_{t:k}(\phi_1, P_t) \right\}, 
\quad \text{where } k \geq t \text{ is the min integer s.t. } f_k(\phi_2, P_t) > 0.
\label{eq14}
\end{equation}
\begin{equation}
f_t(\phi_1 \, S \, \phi_2, P_t) = \min^\gamma \left\{ f_{k:T-1}(\phi_1, P_t) \right\}, \quad \text{where } k \geq t \text{ is the min integer s.t. } f_k(\phi_2, P_t) > 0.
\label{eq15}
\end{equation}

In Eq.\eqref{eq14}, \( \min_\gamma \left\{ f_{t:k}(\phi_1, P_t) \right\} \) is positive (True) if and only if all elements of \( \left\{ f_{t:k}(\phi_1, P_t) \right\} \) are positive, meaning \( \phi_1 \) remains True from time \( t \) to time \( k \). This aligns with the semantics of \( W \). Similarly, \( \min_\gamma \) is used for \( S \) in Eq.\eqref{eq15} for the same reason.

The definitions in Eq.\eqref{eq11}-\eqref{eq15} ensure the logical soundness of \( \Phi \), allowing \( f_t(\phi, P_t) > 0 \) to serve as an optimization objective to enforce the constraints specified by \( \phi \) on a task model. Formally, as \( \gamma \to \infty \), the approximated evaluation (due to \( \min_\gamma \)) becomes exact, and the following theorem holds by construction:

Theorem 1. (Soundness) For \( \phi \in \Phi \), as \( \gamma \to \infty \): if \( f_t(\phi, P_t) > 0 \), then \( \phi \) is satisfied by \( P_t \) at time \( t \).

Essentially, the evaluation process propagates network recognitions from atomic propositions to logic operators and finally to the formula. Equivalently, we can represent a formula as a directed acyclic graph (DAG), where each leaf node is labeled with True, False, or phase P; and each internal node is labeled with logic operators. The edges of the DAG point from child nodes to their parents, along which the truth value propagates following Eqn.\eqref{eq11}-\eqref{eq15}. Fig.\ref{CPKC}(b) and (c) illustrates how a constraint “a person cannot Incision before Vessel\_treatment” is converted to a DAG and evaluated.

\subsection{Loss Function}
During the training process, we aim for the output \( P_t \) to be constrained by both the ground truth \( Y_0 \) and the phase logic constraint rules described by \( \phi \). The constraints from ground truth are enforced by a task loss \(L_{Task}\), which consists of three loss functions.

\textbf{Cross-Entropy Loss.} The first loss is the standard cross-entropy loss for classification, which minimizes the negative log-likelihood of the ground truth phase class for each frame:
\begin{equation}
\mathcal{L}_{\text{CE}} = \frac{1}{T \cdot C} \sum_{i=1}^{T} \sum_{c=1}^{C} -Y_{0,i,c} \log P_{t,i,c}.
\label{eq16}
\end{equation}
where \( i \) is the frame index, \( c \) is the class index, \( Y_{0,i,c} \) is the ground truth label, and \( P_{t,i,c} \) is the predicted probability for class \( c \) at frame \( i \).

\textbf{Temporal Smoothness Loss.} To encourage local consistency along the temporal dimension, the second loss is computed as the mean squared error of the log-likelihoods between adjacent frames \cite{farha2019ms, li2022bridge}:
\begin{equation}
\mathcal{L}_{\text{SMO}} = \frac{1}{(T-1) \cdot C} \sum_{i=1}^{T-1} \sum_{c=1}^{C} (\log P_{t,i,c} - \log P_{t,i+1,c})^2
\label{eq17}
\end{equation}
Note that \( \mathcal{L}_{\text{SMO}} \) is clipped to avoid the influence of outlier values \cite{farha2019ms}.

\textbf{Boundary Alignment Loss.} Accurate detection of phase boundaries is crucial for surgical phase recognition. The third loss aligns the phase boundaries in the denoised sequence \( P_t \) with those in the ground truth sequence \( Y_0 \). To achieve this, boundary probabilities are derived from both \( P_t \) and \( Y_0 \). First, a ground truth boundary sequence \( B \in \{0, 1\}^{T-1} \) is derived from \( Y_0 \), where \( B_i = \mathbf{1}(Y_{0,i} \neq Y_{0,i+1}) \). Since phase transitions usually happen gradually, this sequence is smoothed using a Gaussian filter to obtain a soft version \( \bar{B} = \omega(B) \). Then, the boundary probabilities in the denoised sequence \( P_t \) are computed as the dot product of phase probabilities from neighboring frames in \(1 - P_{s,i} \cdot P_{s,i+1}\). Finally, the boundaries from the two sources are aligned using a binary cross-entropy loss:
\begin{equation}
\mathcal{L}_{\text{BD}} = \frac{1}{T-1} \sum_{i=1}^{T-1} \left[ -\bar{B}_i \log(1 - P_{s,i} \cdot P_{s,i+1}) - (1 - \bar{B}_i) \log(P_{s,i} \cdot P_{s,i+1}) \right].
\label{eq18}
\end{equation}

For phase logic constraint rules, we treat \( P_t \) as an assignment to \( \phi \), and from Theorem 1, we know that \( f_t(\phi, P_t) > 0 \) if \( P_t \) satisfies \( \phi \) starting from time \( t \). Therefore, we can minimize the following objective:
\begin{equation}
\mathcal{L}_{\text{PL}} = \log(1 + e^{-x})(f_0(\phi, P_t)).
\label{eq19}
\end{equation}
In our experiments, we set \( t = 0 \) in \( f_t(\phi, P_t) \) because we require the predictions to satisfy the constraints starting from the first frame. However, \( t \) can be set to different values to flexibly apply constraints at various temporal locations, depending on the specific requirements of the task.

The final training loss is a combination of task loss \(L_{Task}\) and phase logic loss \(L_{PL}\) at a randomly selected diffusion step:
\begin{equation}
\mathcal{L}_{\text{sum}} = \lambda_{CE}\mathcal{L}_{\text{CE}} + \lambda_{SMO}\mathcal{L}_{\text{SMO}} + \lambda_{BD}\mathcal{L}_{\text{BD}} + \lambda_{PL}\mathcal{L}_{\text{PL}}, \quad t \in \{1, 2, \dots, T\}.
\label{eq20}
\end{equation}
where $\lambda_{CE}$, $\lambda_{SMO}$, $\lambda_{BD}$, and $\lambda_{PL}$ are the weight of the loss fuctions $\mathcal{L}_{\text{CE}}$, $\mathcal{L}_{\text{SMO}}$, $\mathcal{L}_{\text{BD}}$, and $\mathcal{L}_{\text{PL}}$, respectively. In this paper $\lambda_{CE}$, $\lambda_{SMO}$, $\lambda_{BD}$, and $\lambda_{PL}$ are set to 0.5, 0.025, 0.1, and 0.1, respectively. 

\section{Experiments and Results}
\label{sec:experiments}
This section first details our experimental framework, including dataset specifications, evaluation metrics, and implementation pdetails. We then validate our proposed method through: (1) comparative analysis with state-of-the-art methods, (2) systematic ablation studies, and (3) multicenter dataset validation, collectively demonstrating its superior performance and strong generalization capability across diverse clinical settings.
\subsection{Experiments design}
\begin{figure*}[!t]
	\centerline{\includegraphics[width=\textwidth]{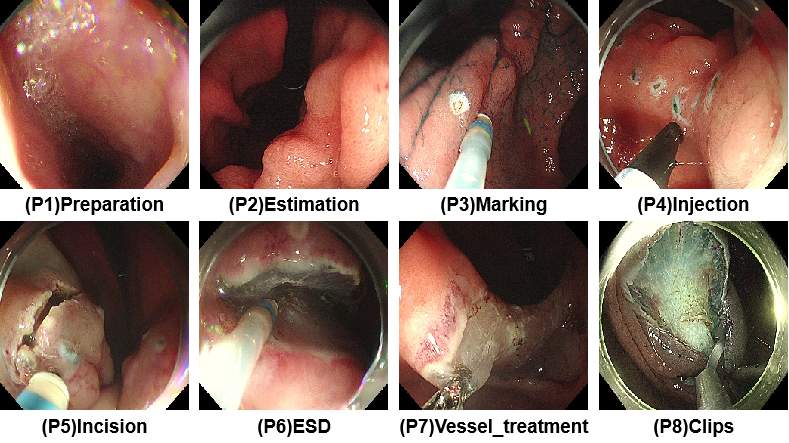}}
	\caption{Illustration of 8 surgical phases (P1-P8) annotated in the ESD820 dataset.}
	\label{fig7}
\end{figure*}

\begin{table}[ht]
\caption{Distribution of annotations in the ESD820 dataset for phase recognition.}
\label{tab10}
\centering
\setlength{\tabcolsep}{6pt}
\begin{tabular}{c|c|c|c}
\toprule
 & Train & Val & Test \\
\hline
Preparation & 200713 & 23465 & 60472 \\
\hline
Estimation & 98596 & 13070 & 27062 \\ 
\hline
Marking & 39173 & 4881 & 11472 \\
\hline
Injection & 95283 & 12021 & 27254 \\
\hline
Incision & 146174 & 16543 & 38662 \\
\hline
ESD & 441097 & 45290 & 118681 \\
\hline
Vessel\_treatment & 92868 & 11816 & 23270 \\
\hline
Clips & 26081 & 5141 & 6670 \\
\hline
Total & 1139985 & 112227 & 313543 \\
\bottomrule
\end{tabular}
\end{table}
\textbf{Dataset.} 
We conducted comprehensive experiments on two large-scale surgical video datasets, namely ESD820 \cite{zhang2024sprmamba, chen2025renji} and Cholec80 \cite{twinanda2016endonet}, endoscopic submucosal dissection and capturing cholecystectomy, respectively. ESD820 \cite{zhang2024sprmamba, chen2025renji} consists of 820 ESD surgical videos manually annotated into 8 phases: Preparation, Estimation, Marking, Injection, Incision, ESD, Vessel\-treatment, and Clips(portion). Samples are shown in Fig.\ref{fig7}. Surgical videos are recorded at 50 fps of a resolution of 1920×1080 pixels with an average video duration of 32 min. We split the dataset into 586 videos for training, 74 videos for validation, and 160 videos for testing. Detailed statistics for each phase are shown in Table \ref{tab10}. Overall, a total of 1139985 frames, 112227, and 313543 frames were annotated for training, validation, and verification, respectively. Cholec80 \cite{twinanda2016endonet} comprises 80 cholecystectomy videos performed by 13 surgeons. These videos were captured at 25 frames per second, with resolutions of either 1920×1080 or 854×480 pixels. Each frame is annotated with both surgical phases (7 possible classes) and instrument presence. For our study, we exclusively utilized the phase annotations. Following established experimental protocols from prior works \cite{twinanda2016endonet,jin2017sv,jin2021temporal}, we adopted a standardized data split: the first 40 videos for training and the remaining 40 videos for testing. To maintain consistency with prior studies \cite{twinanda2016endonet, jin2017sv,jin2021temporal, zhang2024sprmamba}, both datasets were processed by downsampling to 1 frame per second and adjusting the resolution to 250×250 pixels.  

\textbf{Evalutation Metrics.}
Our performance evaluation employs four well-established metrics - Accuracy, Precision, Recall, and Jaccard Index. Accuracy quantifies frame-wise classification correctness across the entire video, while Precision, Recall, and Jaccard Index address class imbalance through phase-level computation followed by macro-averaging, ensuring equitable evaluation of all surgical phases regardless of their temporal prevalence. For Cholec80 benchmarking, we strictly follow the relaxed evaluation protocol \cite{twinanda2016single} adopted by recent state-of-the-art methods \cite{twinanda2016endonet,jin2017sv,jin2020multi,jin2021temporal}, where predictions within 10 seconds of ground truth annotations are considered valid matches. This standardized approach maintains comparability with existing research while providing a comprehensive assessment of both frame-level accuracy and phase-wise recognition quality. All other experiments utilize conventional exact-match metrics unless otherwise specified.

\textbf{Implementation Details.}
We implement our method using PyTorch on an NVIDIA RTX 4090 GPU workstation. To train the STFE unit, Swin-B is initialized with ImageNet-22K pre-trained weights \cite{he2016deep} and fine-tuned on our surgical dataset with a batch size of 96. During training, we apply standard data augmentation including random 224×224 cropping, horizontal/vertical flipping, and mirroring to enhance generalization. We trained the STFE with AdamW for 200 epochs, with a 5-epoch warmup \cite{loshchilov2017decoupled} and cosine annealed decay. For the diffusion process, our diffusion unit utilizes dataset-specific encoder architectures, featuring 14 layers with 256 feature maps for ESD820 and 12 layers with 256 feature maps for Cholec80. To optimize computational efficiency during the iterative denoising process, we design a lightweight decoder structure consisting of 8 layers and 128 feature maps for both datasets. The conditioning features \(F_T\) are extracted from encoder layer 11 in ESD820 and layer 9 in Cholec80. The entire network undergoes end-to-end training using the Adam optimizer \cite{loshchilov2017decoupled} with a fixed batch size of 4 and a learning rate of 5e-4 for both datasets. Beyond the main decoder objective \(L_{sum}\), we implement auxiliary supervision via three complementary losses (\(L_{CE}, L_{SMO}, L_{PL}\) applied to the encoder's prediction head. The total steps are set as S=1000, while inference utilizes an 8-step sampling strategy with steps skipping \cite{song2020denoising}. Following Eq.\eqref{eq5} and \eqref{eq7}, phase sequences are normalized to the range [-1, 1] for both noise addition and removal operations. Notably, our framework processes all video frames and predicts all surgical phases simultaneously, eschewing auto-regressive approaches during both training and inference phases.

\subsection{Comparison with state-of-the-art methods}
\subsubsection{Result on the ESD820 Dataset}
On the ESD820 dataset, we compared our proposed method with several state-of-the-art approaches, including: (1) the method by Furube et al. \cite{furube2024automated}, which fine-tunes ResNet-50 \cite{he2016deep} as a feature extractor followed by MS-TCN for hierarchical prediction refinement; (2) the intelligent surgical workflow recognition suite for ESD proposed by Cao et al. \cite{cao2023intelligent}, based on Trans-SVNet \cite{gao2021trans}; (3) SV-RCNet \cite{jin2017sv} and SAHC \cite{ding2022exploring}, two SOTA methods for cholecystectomy surgical phase recognition; and (4) ASFormer \cite{asformer}, a SOTA transformer-based approach for action segmentation. 
\begin{table}[ht]
\caption{Comparison with the SOTA methods on the ESD820 dataset. Note that the bold text indicates the current optimal result.}
\label{tab1}
\centering
\setlength{\tabcolsep}{3pt}
\resizebox{\textwidth}{!}{
\begin{tabular}{c|c|c|c|c|c}
\toprule
Method & Accuracy(\%) & Precision(\%) & Recall(\%) & Jaccard(\%) & FPS(frames/s)\\
\hline
Swin-B \cite{2021Swin} & 80.60$\pm$11.36 & 79.90$\pm$15.79 & 80.06$\pm$16.15 & 67.27$\pm$15.42 & 233\\
\hline
SV-RCNet \cite{jin2017sv} & 82.46$\pm$14.16 & 84.97$\pm$17.13 & 83.62$\pm$18.11 & 72.84$\pm$20.09 & 30\\
\hline
SAHC \cite{ding2022exploring} & 85.01$\pm$11.92 & 83.22$\pm$17.00 & 86.44$\pm$17.27 & 74.03$\pm$18.93 & 217\\
\hline
Furube et al. \cite{furube2024automated} & 85.63$\pm$11.36 & 84.43$\pm$16.24 & 86.48$\pm$17.76 & 74.94$\pm$19.01 & 226\\
\hline
AI-Endo \cite{cao2023intelligent} & 85.24$\pm$11.10 & 84.45$\pm$15.02 & 86.05$\pm$17.04 & 74.72$\pm$17.92 & 220\\
\hline
SPRMamba \cite{zhang2024sprmamba} & 86.35$\pm$12.59 & 86.04$\pm$16.35 & 88.80$\pm$16.62 & 77.90$\pm$18.11 & 178\\
\hline
ASFormer \cite{asformer} & 85.54$\pm$12.86 & 85.36$\pm$16.61 & 87.39$\pm$16.79 & 76.33$\pm$18.59 & 187\\
\hline
\textbf{STFE (Ours)} & 87.28$\pm$9.60 & 87.76$\pm$12.87 & 88.53$\pm$12.49 & 79.05$\pm$15.16 & 224\\
\hline
\textbf{CPKD (Ours)} & \textbf{88.88$\pm$10.95} & \textbf{89.59$\pm$14.73} & \textbf{91.02$\pm$14.63} & \textbf{82.59$\pm$16.36} & 218\\
\bottomrule
\end{tabular}
}
\end{table}

As shown in Table \ref{tab1}, our experimental results reveal significant performance improvements through the proposed architecture. The standalone STFE unit achieves baseline performance with 87.28\% accuracy, 87.76\% precision, 88.53\% recall, and 79.05\% Jaccard index. With the integration of both Diffusion and CPKC units, the present method demonstrates substantial gains, reaching 88.88\% accuracy (+1.6\% absolute improvement), 88.59\% precision, 91.02\% recall, and 82.59\% Jaccard score. When compared against state-of-the-art approaches, our method establishes new benchmarks across all metrics, surpassing the previous best results by margins of 2.53\% (Accuracy), 3.55\% (Precision), 2.22\% (Recall), and 4.69\% (Jaccard). Furthermore, to demonstrate the computational efficiency of the proposed method, we also compared our method with the state-of-the-art (SOTA) method in terms of Frames Per Second (FPS), as shown in Table \ref{tab1}. Our method, equipped with a lightweight decoder, significantly outperforms ASFormer in terms of FPS.
\begin{figure*}[!t]
	\centerline{\includegraphics[width=\textwidth]{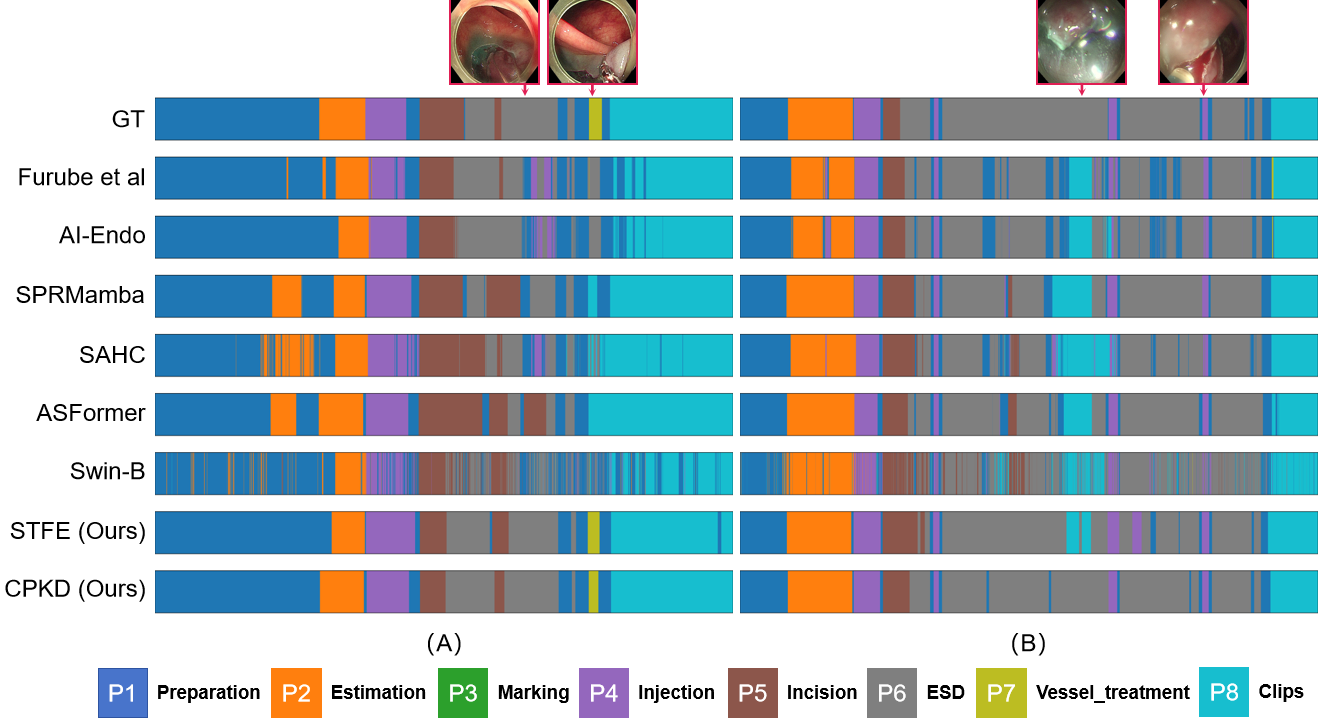}}
	\caption{Qualitative comparisons with baseline and the SOTA methods on ESD820 dataset. The first row highlights critical frames (indicated by red arrows) where existing SOTA methods fail to classify phases correctly, particularly in ambiguous scenarios such as subtle instrument-tissue interactions or occluded views. Subsequent rows illustrate the ground-truth (GT) phases (P1–P8) alongside predictions from competing methods.}
	\label{fig4}
\end{figure*}

To illustrate the performance of our approach compared to the SOTA, in Fig.\ref{fig4} we present a qualitative comparison of two examples drawn from the ESD820 test datasets. As observed in Fig.\ref{fig4}(A), for some ambiguous frames shown (first row), SOTA methods were unable to effectively classify the correct phase. Specifically, SPRMamba and ASFormer predict P7 as P8, SAHC predicts P6 and P7 alternately, and other SOTA methods predict P6. Additionally, Fig.\ref{fig4}(B) shows that our method exhibits superior consistency with GT in capturing longer continuous phases, whereas SOTA methods incorrectly fragment single phases into shorter segments. This consistency is further emphasized when analyzing the impact of individual units: using only the STFE unit can lead to inaccurate segmentation boundaries and misclassified phases, likely due to insufficient prior modeling. Adding the other two units significantly reduces these errors, achieving smoother and more accurate phase segmentation.

\subsubsection{Result on the Cholec80 Dataset}
Our comprehensive evaluation on the Cholec80 benchmark compares the proposed approach against nine established methods: TeCNO \cite{czempiel2020tecno}, Trans-SVNet \cite{gao2021trans}, SKiT \cite{liu2023skit}, LoViT \cite{liu2025lovit}, SPRMamba \cite{zhang2024sprmamba}, SR-Mamba\cite{cao2024sr}, Surgformer \cite{yang2024surgformer}, and BNpitfalls \cite{rivoir2024pitfalls}. For experimental consistency, we re-implemented TeCNO and Trans-SVNet using the authors' provided model weights, while directly reporting published results for other comparative methods from their original papers. 
\begin{table}[ht]
\caption{Comparison with the SOTA methods on the Cholec80 dataset.}
\label{tab2}
\centering
\setlength{\tabcolsep}{3pt}
\resizebox{\textwidth}{!}{
\begin{tabular}{c|c|c|c|c|c}
\toprule
Method & Relaxed metric & Accuracy(\%) & Precision(\%) & Recall(\%) & Jaccard(\%) \\
\hline
TeCNO \cite{czempiel2020tecno} & \checkmark & 90.17$\pm$6.91 & 88.13$\pm$5.29 & 87.07$\pm$6.62 & 76.01$\pm$7.78 \\
\hline
Trans-SVNet \cite{gao2021trans} & \checkmark & 91.54$\pm$6.76 & 91.15$\pm$4.12 & 89.47$\pm$6.32 & 79.89$\pm$6.44 \\
\hline
SKiT \cite{liu2023skit} & \checkmark & 93.40$\pm$5.20 & 90.90 & 91.80 & 82.60 \\
\hline
LoViT \cite{liu2025lovit} & \checkmark & 92.40$\pm$6.30 & 89.90$\pm$6.10 & 90.60$\pm$4.40 & 81.20$\pm$9.10 \\
\hline
SPRMamba \cite{zhang2024sprmamba} & \checkmark & 93.60$\pm$5.27 & 90.64$\pm$5.09 & 92.43$\pm$5.09 & 83.35$\pm$6.37 \\
\hline
SR-Mamba \cite{cao2024sr} & \checkmark & 92.60$\pm$8.60 & 90.30$\pm$5.20 & 90.60$\pm$7.20 & 81.50$\pm$8.60 \\
\hline
Surgformer \cite{yang2024surgformer} & \checkmark & 93.40$\pm$6.40 & \textbf{91.90$\pm$4.70} & 92.10$\pm$5.80 & 84.10$\pm$8.00 \\
\hline
BNpitfalls \cite{rivoir2024pitfalls} & \checkmark & 93.50$\pm$6.50 & 91.60$\pm$5.00 & 91.40$\pm$9.30 & 82.90$\pm$10.10 \\
\hline
\textbf{STFE+Diffusion (Ours)} & \checkmark & \textbf{94.15$\pm$6.24} & 91.12$\pm$4.73 & \textbf{94.12$\pm$3.29} & \textbf{85.61$\pm$5.21} \\
\hline
TeCNO \cite{czempiel2020tecno} &  & 89.35$\pm$6.70 & 83.24$\pm$7.21 & 81.29$\pm$6.61 & 70.08$\pm$9.08 \\
\hline
Trans-SVNet \cite{gao2021trans} &  & 90.27$\pm$6.48 & 85.23$\pm$6.97 & 82.92$\pm$6.77 & 72.42$\pm$8.92 \\
\hline
SKiT \cite{liu2023skit} &  & 92.50$\pm$5.10 & 84.60 & 88.50 & 76.70 \\
\hline
LoViT \cite{liu2025lovit} &  & 91.50$\pm$6.10 & 83.10$\pm$9.30 & 86.50$\pm$5.50 & 74.20$\pm$11.30 \\
\hline
SPRMamba \cite{zhang2024sprmamba} &  & 93.12$\pm$4.58 & \textbf{89.26$\pm$6.69} & 90.12$\pm$5.61 & 81.43$\pm$6.90 \\
\hline
Surgformer \cite{yang2024surgformer} & & 92.40$\pm$6.40 & 87.90$\pm$6.90 & 89.30$\pm$7.80 & 79.90$\pm$10.20 \\
\hline
\textbf{STFE+Diffusion (Ours)} &  & \textbf{93.47$\pm$6.29} & 89.06$\pm$4.73 & \textbf{91.29$\pm$7.01} & \textbf{82.24$\pm$9.24} \\
\bottomrule
\end{tabular}
}
\end{table}

As shown in Table \ref{tab2}, our method establishes new state-of-the-art performance across multiple metrics, achieving the highest accuracy (94.15\%±6.24 vs 93.60\%±5.27 for SPRMamba), recall (94.12\%±3.29 vs 92.43\%±6.09 for SPRMamba), and Jaccard index (85.61\%±5.21 vs 84.10\%±8.00 for Surgformer). While Surgformer shows a marginal advantage in precision (91.90\%±4.70 vs our 91.12\%±4.73), our approach maintains strong competitiveness in this metric while delivering more balanced and consistent performance across all evaluation criteria, as evidenced by lower standard deviations in most metrics. Particularly noteworthy is our method's significant improvement in recall (1.69\% absolute gain over SPRMamba) and Jaccard index (1.51\% over Surgformer), which demonstrates enhanced capability in capturing complete phase segments and maintaining temporal consistency. These quantitative results validate the effectiveness of integrating diffusion models with surgical phase recognition, especially in handling complex temporal dependencies and boundary ambiguities inherent in endoscopic videos.

\begin{figure*}[!t]
	\centerline{\includegraphics[width=\textwidth]{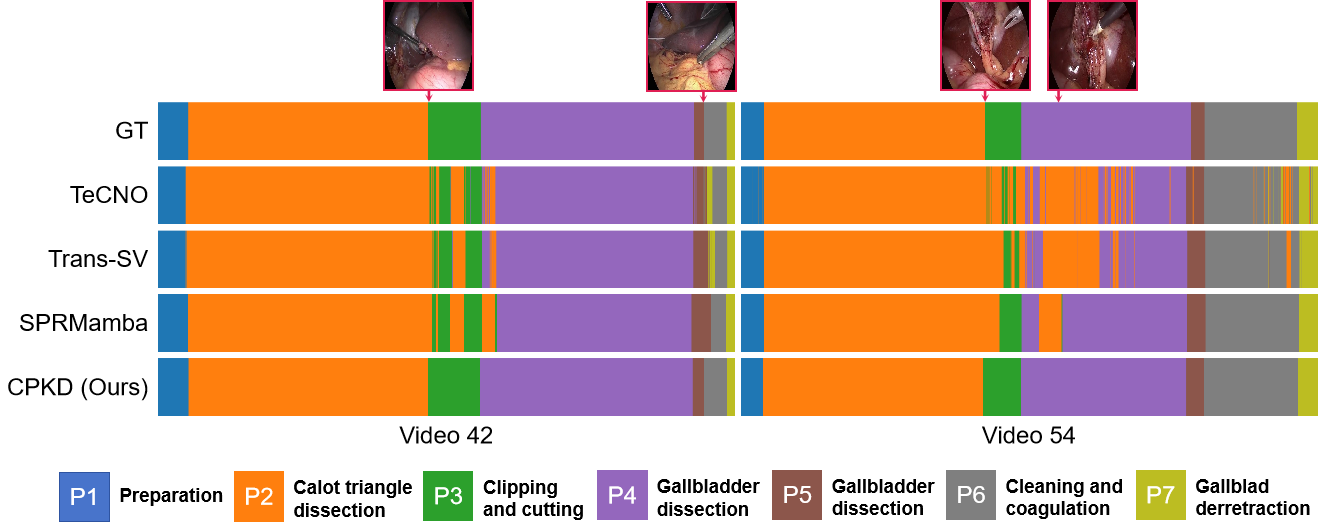}}
	\caption{Qualitative comparisons with baseline and the SOTA methods on Cholec80 dataset. The first row shows some images from the video corresponding to the moments indicated by the red arrows, indicating examples of errors in SOTA methods other than ours. The following five rows represent the ground truth and the corresponding phase results recognized by different methods. P1 to P7 indicate the phase label.}
	\label{fig5}
\end{figure*}

Moreover, Fig.\ref{fig5} presents a qualitative comparison between our method and other state-of-the-art approaches (TeCNO, Trans-SVNet, and SPRMamba) on two examples from the Cholec80 dataset. Visual inspection reveals that our method aligns more closely with ground-truth labels, demonstrating superior phase recognition accuracy. For instance, in transition phases (e.g., P2 to P3), where other methods exhibit misclassifications or temporal inconsistencies, our predictions maintain higher temporal coherence and boundary precision. Notably, our method exhibits high sensitivity to phases with shorter durations and better captures fine-grained phase transitions. Compared to GT, it can more accurately recognize the beginning and end of these phases. In contrast, the SOTA methods suffers from missing or insufficient recognition during these critical transient phases.

\subsection{Ablation Study}
We conducted a series of ablation experiments on the ESD820 dataset to validate the effectiveness of each component and parameter setting in our proposed method on the model.

\subsubsection{Effectiveness of Conditional Masking Strategy} 
\begin{table}[ht]
\caption{Ablation study on the conditional masking strategy.}
\label{tab4}
\centering
\setlength{\tabcolsep}{3pt}
\resizebox{\textwidth}{!}{
\begin{tabular}{c|c|c|c|c|c|c|c}
\toprule
N & G & T & R & Accuracy(\%) & Precision(\%) & Recall(\%) & Jaccard(\%) \\
\hline
\checkmark &  &  &  & 88.53$\pm$9.66 & 88.24$\pm$13.98 & 90.93$\pm$14.33 & 81.31$\pm$16.22 \\
\hline
\checkmark & \checkmark &  &  & 88.22$\pm$11.23 & 88.77$\pm$14.91 & 90.49$\pm$14.82 & 81.52$\pm$16.69 \\
\hline
\checkmark &  & \checkmark &  & \textbf{88.91$\pm$9.46} & 88.81$\pm$14.11 & 90.91$\pm$14.21 & 81.82$\pm$15.82 \\
\hline
\checkmark &  &  & \checkmark & 88.29$\pm$11.13 & 88.92$\pm$14.38 & 90.89$\pm$14.30 & 81.95$\pm$16.34 \\
\hline
\checkmark & \checkmark & \checkmark &  & 88.34$\pm$11.15 & 88.93$\pm$14.62 & \textbf{91.05$\pm$14.54} & 82.12$\pm$16.53 \\
\hline
\checkmark & \checkmark &  & \checkmark & 88.27$\pm$11.24 & 
88.86$\pm$15.18 & 90.81$\pm$15.05 & 81.85$\pm$16.67 \\
\hline
\checkmark &  & \checkmark & \checkmark & 88.21$\pm$11.33 & 88.65$\pm$14.67 & 90.92$\pm$13.63 & 81.71$\pm$16.52 \\
\hline
\checkmark & \checkmark & \checkmark & \checkmark & 88.88$\pm$10.95 & \textbf{89.59$\pm$14.73} & 91.02$\pm$14.63 & \textbf{82.59$\pm$16.36} \\
\bottomrule
\end{tabular}
}
\end{table}
Our ablation study systematically evaluates different combinations of condition masking schemes. The results demonstrate that the comprehensive integration of all three masking strategies achieves optimal performance. Notably, the phase relation masking strategy is especially useful among the three strategies, with a increase result from 81.31 Jaccard to 82.12 Jaccard when the phase relation masking strategy is added. These results indicate that explicitly capturing ordinal phase relationships through conditional masking strategy offers vital contextual information, effectively complementing both global scene understanding and transition-aware features in surgical phase recognition.

\subsubsection{Effectiveness of Conditioning Features}
\begin{table}[ht]
\caption{Ablation study on the conditioning features.}
\label{tab5}
\centering
\setlength{\tabcolsep}{3pt}
\resizebox{\textwidth}{!}{
\begin{tabular}{c|c|c|c|c}
\toprule
Features & Accuracy(\%) & Precision(\%) & Recall(\%) & Jaccard(\%) \\
\hline
Spatial Feature & 86.39$\pm$15.27 & 88.19$\pm$17.77 & 89.04$\pm$17.75 & 79.80$\pm$18.72 \\
\hline
\( R_\varphi \) Layer 7 & 87.20$\pm$14.90 & 88.89$\pm$16.93 & 90.55$\pm$16.74 & 81.65$\pm$18.23 \\
\hline
\( R_\varphi \) Layer 9 & 88.25$\pm$12.51 & 89.12$\pm$15.51 & 90.90$\pm$15.82 & 82.16$\pm$17.38 \\
\hline
\( R_\varphi \) Layer 11 & \textbf{88.88$\pm$10.95} & \textbf{89.59$\pm$14.73} & \textbf{91.02$\pm$14.63} & \textbf{82.59$\pm$16.36} \\
\hline
\( R_\varphi \) Layer 7,9,11 & 87.99$\pm$11.44 & 89.02$\pm$15.49 & 90.57$\pm$15.13 & 81.83$\pm$17.10 \\
\hline
\( R_\varphi \) Prediction & 87.59$\pm$13.49 & 88.95$\pm$16.50 & 90.48$\pm$16.97 & 81.64$\pm$18.17 \\
\bottomrule
\end{tabular}
}
\end{table}
For the generation condition, the intermediate features from the encoder $R_{\varphi}$ are used in Eq.\eqref{eq6}. In Table \ref{tab5}, we ablate over different choices for the conditioning features, including the spatial feature and the features from different layers of the encoder $R_{\varphi}$. Performance drops remarkably when using the spatial feature F as the condition (86.39 Acc vs. 88.88 Avg), suggesting the necessity of an encoder. On the other hand, the choice of encoder layer substantially impacts model effectiveness, where deeper layers (particularly Layer 11) consistently achieve superior results across all metrics (Peak Jaccard: 82.59\%±16.36). This layer-depth performance correlation suggests that higher-level semantic features - which better capture procedural context and instrument-tissue interactions - are more discriminative for phase recognition than low-level visual features. Interestingly, combining multiple layers (Layers 7,9,11) underperforms using Layer 11 alone, indicating potential feature redundancy in multi-layer fusion.

\subsubsection{Effectiveness of Training Task Loss}
\begin{table}[ht]
\caption{Ablation study on the task loss functions.}
\label{tab12}
\centering
\setlength{\tabcolsep}{3pt}
\resizebox{\textwidth}{!}{
\begin{tabular}{c|c|c|c|c|c|c}
\toprule
\(\mathcal{L}_{\text{CE}}\) & \(\mathcal{L}_{\text{SMO}}\) & \(\mathcal{L}_{\text{BD}}\) & Accuracy(\%) & Precision(\%) & Recall(\%) & Jaccard(\%) \\
\hline
\checkmark &  &  & 87.87$\pm$11.51 & 88.43$\pm$14.95 & 90.40$\pm$15.58 & 81.14$\pm$17.08 \\
\hline
\checkmark & \checkmark &  & 88.39$\pm$10.85 & 88.74$\pm$14.28 & \textbf{91.27$\pm$14.31} & 82.05$\pm$15.95 \\
\hline
\checkmark &  & \checkmark & 87.45$\pm$12.36 & 88.48$\pm$15.03 & 90.09$\pm$15.95 & 80.97$\pm$17.28 \\
\hline
\checkmark & \checkmark & \checkmark & \textbf{88.88$\pm$10.95} & \textbf{89.59$\pm$14.73} & 91.02$\pm$14.63 & \textbf{82.59$\pm$16.36} \\
\bottomrule
\end{tabular}
}
\end{table}
In Table \ref{tab12}, we systematically evaluates the effect of task loss components to model performance. Four configurations are compared: (1) \( \mathcal{L}_{\text{CE}} \) alone, (2) \( \mathcal{L}_{\text{CE}} + \mathcal{L}_{\text{SMO}} \), (3) \( \mathcal{L}_{\text{CE}} + \mathcal{L}_{\text{BD}} \), and (4) \( \mathcal{L}_{\text{Task}} \). It is found that all loss components are necessary for the best result. Based on \(\mathcal{L}_{\text{CE}}\) and \(\mathcal{L}_{\text{SMO}}\), our proposed boundary alignment \(\mathcal{L}_{\text{BD}}\) brings performance gains in both frame accuracy and temporal continuity.

\subsubsection{Effectiveness of Inference Steps}
\begin{figure*}[!t]
	\centerline{\includegraphics[width=0.8\textwidth]{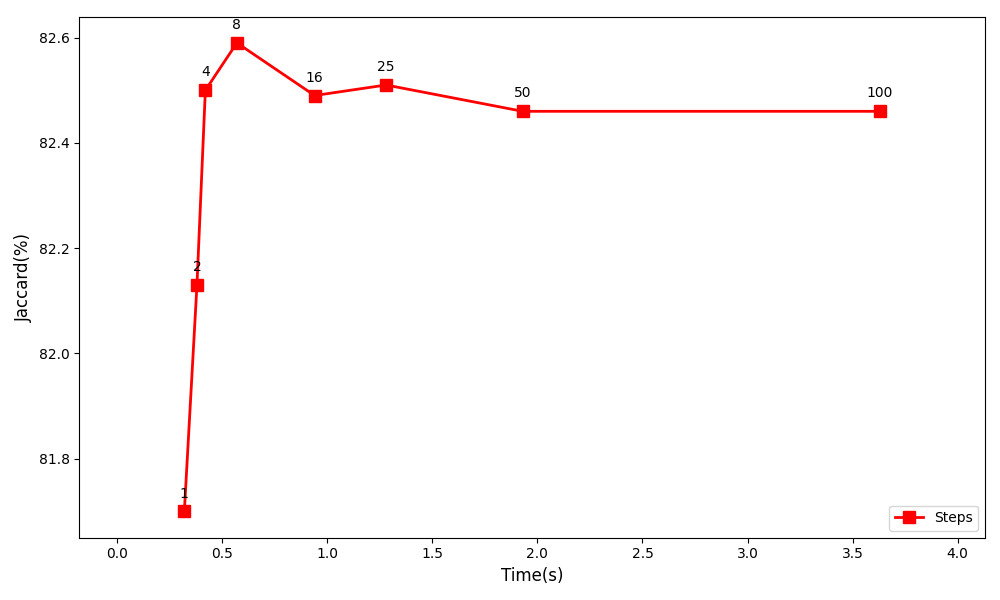}}
	\caption{Inference time visualization of CPKD for different Steps.}
	\label{jaccard-time}
\end{figure*}
Fig.\ref{jaccard-time} presents our ablation study on inference steps selection, revealing several key insights. Performance metrics exhibit steady improvement with increasing step counts, though with diminishing returns beyond 8 steps. Notably, the 8-step configuration achieves optimal balance, delivering peak performance (82.59\% Jaccard) at reasonable computational cost (0.57s). This framework aligns with established findings in multi-stage phase recognition models, confirming our hypothesis about shared refinement mechanisms between diffusion processes and traditional multi-stage approaches. The observed linear time complexity (0.32s to 3.63s across 1-100 steps) motivated our selection of 8 steps as the operational sweet spot between accuracy and efficiency.

\subsubsection{Effectiveness of Clinical Prior Knowledge-Constraint}
\label{sec: CPKC}
\begin{table}[ht]
\caption{Ablation study on the clinical prior knowledge-constraint.}
\label{tab7}
\centering
\setlength{\tabcolsep}{3pt}
\resizebox{\textwidth}{!}{
\begin{tabular}{c|c|c|c|c|c|c}
\toprule
STFE & Diffusion & CPKC & Accuracy(\%) & Precision(\%) & Recall(\%) & Jaccard(\%) \\
\hline
\checkmark &  &  & 87.28$\pm$9.60 & 87.76$\pm$12.87 & 88.53$\pm$12.49 & 79.05$\pm$15.16 \\
\hline
\checkmark & \checkmark &  & 87.81$\pm$12.83 & 88.80$\pm$15.38 & 90.32$\pm$15.68 & 81.41$\pm$16.90 \\
\hline
\checkmark & \checkmark & \checkmark & \textbf{88.88$\pm$10.95} & \textbf{89.59$\pm$14.73} & \textbf{91.02$\pm$14.63} & \textbf{82.59$\pm$16.36} \\
\bottomrule
\end{tabular}
}
\end{table}

\begin{table}[ht]
\caption{Ablation study on the value of $\lambda$}
\label{tab8}
\centering
\setlength{\tabcolsep}{3pt}
\resizebox{\textwidth}{!}{
\begin{tabular}{c|c|c|c|c}
\toprule
 $\lambda$ & Accuracy(\%) & Precision(\%) & Recall(\%) & Jaccard(\%) \\
\hline
0 & 87.81$\pm$12.83 & 88.80$\pm$15.38 & 90.32$\pm$15.68 & 81.41$\pm$16.90 \\
\hline
\textbf{0.1} & \textbf{88.88$\pm$10.95} & \textbf{89.59$\pm$14.73} & 91.02$\pm$14.63 & \textbf{82.59$\pm$16.36} \\
\hline
0.2 & 88.52$\pm$11.22 & 89.05$\pm$15.11 & 90.98$\pm$15.51 & 82.10$\pm$16.97 \\
\hline
0.3 & 88.88$\pm$10.95 & 89.19$\pm$14.58 & \textbf{91.39$\pm$14.46} & 82.57$\pm$16.32 \\
\hline
0.4 & 88.52$\pm$11.20 & 88.72$\pm$14.71 & 91.15$\pm$15.06 & 81.97$\pm$16.87 \\
\hline
0.5 & 87.48$\pm$13.35 & 88.84$\pm$16.22 & 91.04$\pm$16.78 & 82.00$\pm$17.66 \\
\hline
0.7 & 88.43$\pm$11.06 & 89.25$\pm$14.66 & 90.60$\pm$14.86 & 82.02$\pm$16.45 \\
\hline
1.0 & 88.45$\pm$9.60 & 88.60$\pm$14.13 & 90.76$\pm$14.39 & 81.55$\pm$16.03 \\
\bottomrule
\end{tabular}
}
\end{table}

We evaluate the impact of our proposed clinical prior knowledge-constraint unit on surgical phase recognition. As demonstrated in Table \ref{tab7}, the incorporation of clinical prior knowledge constraints during model training consistently improved performance across all metrics on the ESD820 dataset. These results underscore the critical importance of integrating clinical prior knowledge, as it contains the inherent logic relationships between ESD surgical phases. By enforcing these logic constraints, our model achieves a more robust understanding of phase dependencies, thereby significantly reducing logical inconsistencies in the output predictions. In addition, we also investigate the effect of $\lambda$ settings. As shown in Table \ref{tab8}, it can be observed that when $\lambda=0.1$ leads to precision and jaccard get the best results, recall is slightly lower than $\lambda=0.3$. When $\lambda=1.0$, the result is the worst, which may be due to strong logical constraints leading to overfitting of the model. Thus, we choose $\lambda=0.1$.

\begin{figure*}[!t]
	\centerline{\includegraphics[width=\textwidth]{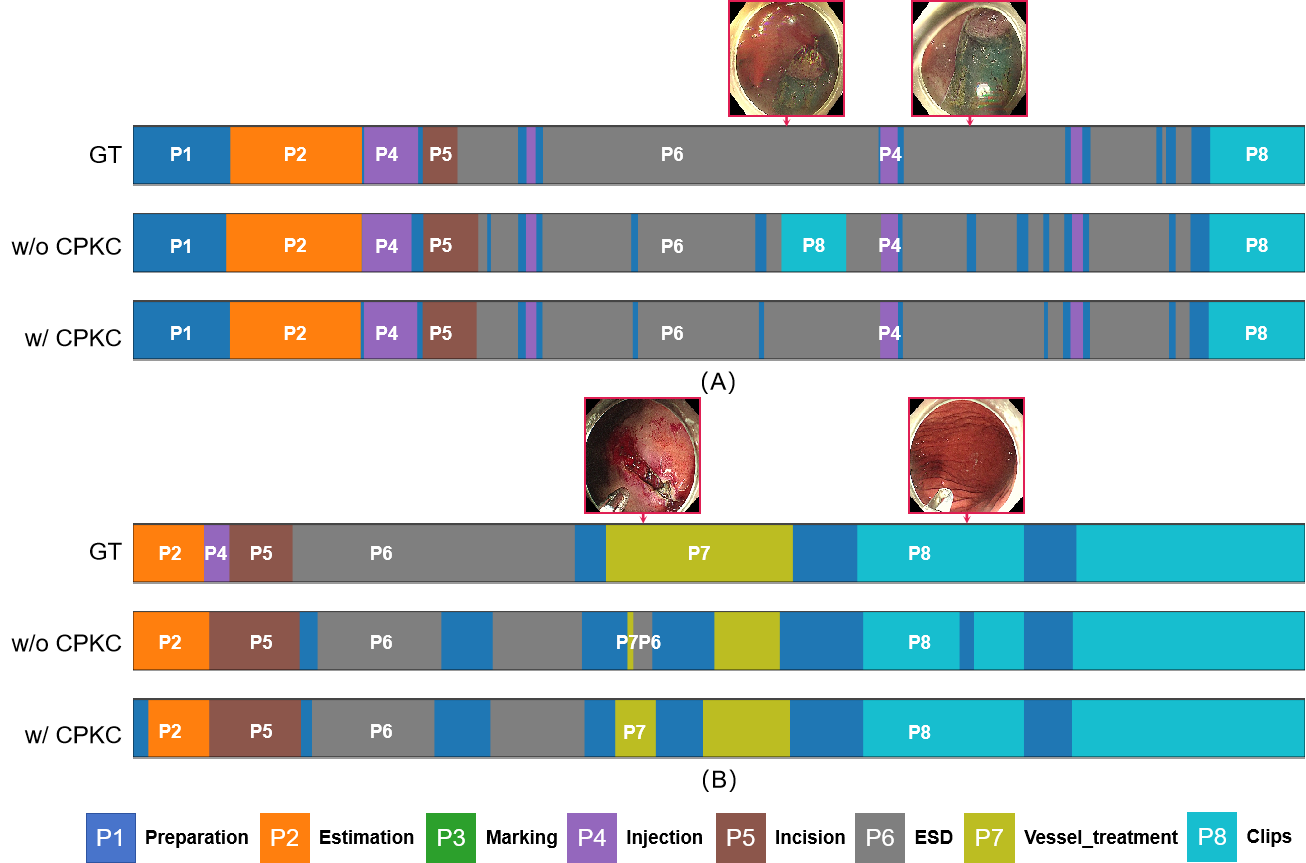}}
	\caption{Qualitative comparisons with between model trained with and without CPKC on ESD820 dataset shows that logical errors are fixed by CPKC. The first row shows some images from the video responding to the moments indicated by the red arrows, indicating examples of logical errors in without CPKC. The following three rows represent the ground truth,  the prediction from without CPKC, and the prediction from our method. P1 to P8 indicate the phase label.}
	\label{fig6}
\end{figure*}

We qualitatively compare the two examples w/ and w/o CPKC units on the ESD820 test dataset in Fig.\ref{fig6}. As shown in Fig.\ref{fig6}(A), w/o CPKC predicts the ESD and the Injection after Clips, which conflicts with the priori knowledge II because injection and ESD cannot be performed after Clips in ESD surgery. Fig.\ref{fig6}(B) incorrectly predicts the ESD operation. It violates priori knowledge II because ESD should not be performed after Vessel\_treatment except for the multifocal case. Both errors were corrected after with CPKC units. Furthermore, we note that correcting individual error fragments improves the quality of the entire output sequence. This is because we are iteratively generating model, and correcting errors fixes their cascading effect on the model state, thus improving the quality of all outputs in the neighborhood.
\subsection{Generalizability of CPKD}
\begin{figure*}[!t]
	\centerline{\includegraphics[width=\textwidth]{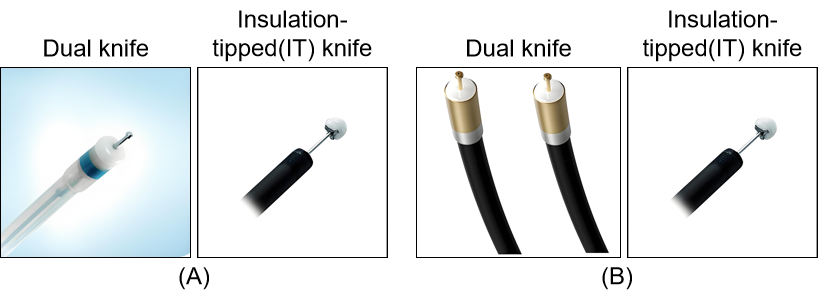}}
	\caption{(A) ESD tools used in Internal dataset. (B) ESD tools used in External dataset.}
	\label{skill}
\end{figure*}
\begin{table}[ht]
\caption{Validation on internal and external datasets with CPKD.}
\label{tab9}
\centering
\setlength{\tabcolsep}{3pt}
\resizebox{\textwidth}{!}{
\begin{tabular}{c|c|c|c|c}
\toprule
Dataset & Accuracy(\%) & Precision(\%) & Recall(\%) & Jaccard(\%) \\
\hline
Internal & 88.88$\pm$10.95 & 89.59$\pm$14.73 & 91.02$\pm$14.63 & 82.59$\pm$16.36 \\
\hline
External & 87.90$\pm$12.38 & 86.57$\pm$15.50 & 87.91$\pm$15.50 & 77.12$\pm$16.49  \\
\bottomrule
\end{tabular}
}
\end{table}
To further validate the generalizability of CPKD, we have collected an external dataset consisting of 95 videos from two different centers, along with their corresponding labeling. We applied CPKD to this external dataset and compared its performance to that of our internal testing dataset. As shown in Table 5, the performance of CPKD on the external dataset is lower than that on the internal test dataset. The performance degradation can be attributed to differences in tool use (Fig.\ref{skill}) and endoscopist levels of ESD experience between the two centers, a common challenge when dealing with multicenter data. Despite this, the results still demonstrate the robustness and generalizability of our method across different datasets and centers.

\section{Disscussion}
Accurate recognition of ESD surgical phases in computer-assisted systems is essential to improve surgical efficiency, reduce patient complications, and provide valuable training material for beginning endoscopists. Previous approaches have accurately recognized surgical phases through multi-stage models. The core idea of the multi-stage model is to correctly model the temporal dependence of surgical phases through an iterative refinement paradigm and to reduce over-recognition errors. However, there are still issues with high computational overhead and inconsistencies in the logic of the recognition results.

To address these issues, we propose a new approach to surgical phase recognition that follows the same philosophy of iterative optimization but employs an essentially new generative approach that incorporates a denoising diffusion model. Furthermore, this is achieved by designing a clinical prior knowledge-constraint unit as well as a generic and novel conditional masking strategy that incorporates the structural prior to the surgical phase into the model training. In this way, we can constrain the training process and guide the model to make more meaningful and logical phase recognition. Specifically, we designed a novel network architecture consisting of three components, namely the STFE unit, the diffusion unit, and the CPKC unit. Each component contributes to the superior performance of the proposed method from both quantitative and qualitative perspectives.
\begin{table}[ht]
\caption{Results on ESD820 using different conditional masking strategies at the inference stage. Note that this is only for analysis purposes since the masking strategies T and R depend on the ground truth.}
\label{tab11}
\centering
\setlength{\tabcolsep}{3pt}
\resizebox{\textwidth}{!}{
\begin{tabular}{c|c|c|c|c}
\toprule
Masking & Accuracy(\%) & Precision(\%) & Recall(\%) & Jaccard(\%) \\
\hline
N & 88.88$\pm$10.95 & 89.59$\pm$14.73 & 91.02$\pm$14.63 & 82.59$\pm$16.36 \\
\hline
G & 28.67$\pm$21.06 & 28.32$\pm$21.78 & 24.19$\pm$21.81 & 13.25$\pm$16.58 \\
\hline
T & 88.91$\pm$10.94 & 89.23$\pm$14.48 & 91.44$\pm$14.44 & 82.64$\pm$16.29 \\
\hline
R & 89.55$\pm$10.75 & 89.71$\pm$14.59 & 92.39$\pm$14.39 & 83.68$\pm$16.19 \\
\bottomrule
\end{tabular}
}
\end{table}

\begin{figure*}[!t]
	\centerline{\includegraphics[width=\textwidth]{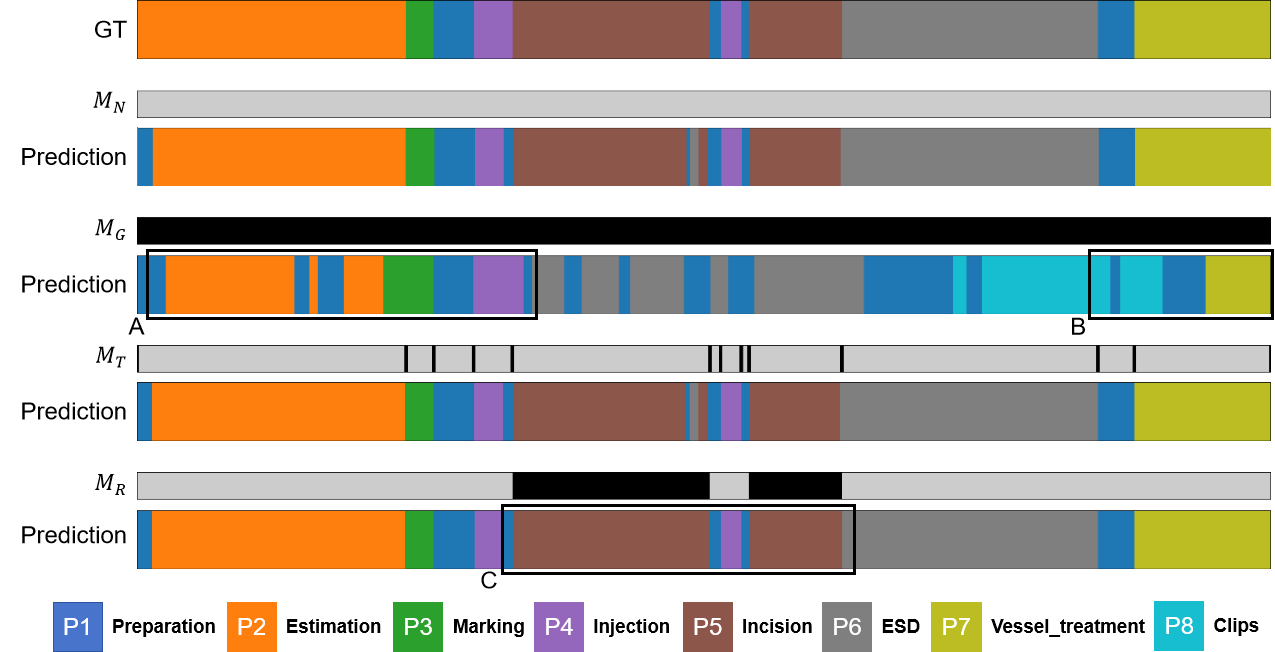}}
	\caption{Visualization of the masks and the corresponding predictions using the masked conditions at inference. In \( M_N \) , \( M_G \) , \( M_T \) , \( M_R \) , masked locations are colored in black.}
	\label{fig8}
\end{figure*}

ESD surgical videos are characterized by domain-specific characteristics and intrinsic logic of each phase, and learning about these priors is crucial for accurately recognizing surgical phases. We first conducted an exploratory experiment to analyze how well the position prior, boundary prior, and relation prior are captured in our model. Recall that the proposed method uses no masking (\(M_N\)) in inference by default, whereas in this experiment we input the condition masking with each conditional masking strategy (\(M_G, M_T, M_R\)) for inference instead. As shown in Table \ref{tab11}, the model even achieves better performance when applying the phase transition masking strategy (\(M_T\)) or the phase relation masking strategy (\(M_R\)), indicating that the boundary prior and the relational prior are well handled. Interestingly, the results using the global masking strategy (\(M_G\)) achieve an Accuracy of 28.67, which is better than random predicting. This reveals that the model learns meaningful associations between phases and temporal locations through the global masking strategy. We further show in Fig.\ref{fig8} the masking strategies and their corresponding phase predictions when each conditional masking strategy is applied to the video during inference. The model generates a generally reasonable sequence of surgical phases when all features are masked by \(M_G\). For example, Estimation and Marking are located at the beginning of the video (Fig.\ref{fig8} A), while Vessel\_treatment and Clips appear at the end (Fig.\ref{fig8} B). With mask \(M_T\), the model is still able to find phase transitions. The missing phase "Incision" masked by \(M_R\) is successfully recovered in Fig.\ref{fig8} C. Then, from the ablation experimental results in section \ref{sec: CPKC}, we observe that the CPKC unit improves the performance and logical reasoning of our method. These analyses demonstrate the capability of our method in prior modeling. Finally, we note that both the CPKC unit and the conditional masking strategy are a simple operation that does not impose an additional burden on the model.

Although our method demonstrates promising performance, there still exist several limitations that warrant further exploration. First, the relatively high computational cost during inference when employing a large number of diffusion steps is a well-known limitation of diffusion models. This issue could be mitigated by improved sampling solvers \cite{lu2022dpm} or diffusion model distillation \cite{2022Progressive}. Second, the advantage of our proposed method is less pronounced on the small-scale Cholec80 dataset compared to its performance on ESD820. We attribute this challenge to the difficulty of learning phase sequence distributions from a limited number of videos, resulting in lower precision. To address this issue in small datasets, a potential solution would be replacing Gaussian noise in the diffusion process with phase-statistics-based perturbations. Lastly, our CPKC unit relies on domain-specific knowledge tailored for ESD surgery, and further experiments are needed to evaluate its generalizability to other surgical procedures that also exhibit prior phase-ordering information. Future research could integrate frame-wise phase sequence generation with segment-level ordered phase lists to achieve diffusion-based surgical phase recognition. Moreover, leveraging our generative framework for unified phase recognition and prediction appears promising, given its inherent suitability for prediction with prior modeling.
\section{Conclusion}
\label{sec:conclusion}
This paper proposes a novel framework for surgical phase recognition in Endoscopic Submucosal Dissection, which generates phase sequences through an iterative denoising process. A flexible conditional masking strategy is designed to jointly leverage positional priors, boundary priors, and relational priors of surgical phases. Furthermore, we incorporate clinical prior knowledge summarized by endoscopic experts into the framework to constrain the training of the surgical phase recognition model, thereby reducing logical errors in the generated phase sequences. We conduct qualitative and quantitative evaluations of the framework and compare it with state-of-the-art methods. Our approach yields superior results and significantly outperforms other methods, opening up new possibilities for surgical phase recognition.  

\section*{Acknowledgment}
This work was supported by the Joint Laboratory of Intelligent Digestive Endoscopy between Shanghai Jiaotong University and Shandong Weigao Hongrui Medical Technology Co., Ltd. and the National Science Foundation of China under Grant 62103263.

\bibliographystyle{unsrt} 
\bibliography{reference}
\end{document}